\documentclass{article} 
\usepackage{iclr2026_conference,times}

\usepackage{amsmath,amsfonts,bm}

\def\eqref#1{equation~\ref{#1}}

\def\1{\bm{1}}

\DeclareMathAlphabet{\mathsfit}{\encodingdefault}{\sfdefault}{m}{sl}
\SetMathAlphabet{\mathsfit}{bold}{\encodingdefault}{\sfdefault}{bx}{n}

\usepackage{hyperref}
\usepackage{url}

\usepackage{duckuments}
\usepackage{xspace}
\usepackage{xcolor}
\usepackage{graphicx}
\usepackage{blindtext}
\usepackage{algorithm}
\usepackage{algpseudocode}
\usepackage{algcompatible}
\usepackage{enumitem}
\usepackage{subcaption}
\usepackage{amsmath}
\usepackage{amsfonts}
\usepackage{mathtools}
\usepackage{booktabs}
\hypersetup{
  linkcolor=BostonGreen,
  citecolor=BostonGreen,
  urlcolor=BostonGreen,
  colorlinks=true,
  filecolor=magenta, 
}
\usepackage{hyperref}       
\usepackage{bm}
\usepackage{wrapfig}
\usepackage{lipsum}
\usepackage{multirow}
\usepackage{cleveref}
\usepackage{tablefootnote}
\usepackage{tikz}
\usepackage{pifont}
\usepackage[table]{xcolor}
\usepackage{makecell}
\newlength\savewidth\newcommand\shline{\noalign{\global\savewidth\arrayrulewidth
  \global\arrayrulewidth 1pt}\hline\noalign{\global\arrayrulewidth\savewidth}}

\usepackage{listings}
\usepackage[endLComment=,italicComments=false]{algpseudocodex}
\usepackage{bbm}

\definecolor{Gray}{gray}{0.9}
\definecolor{Blue9}{rgb}{0.098,0.3,0.9}
\definecolor{Red7}{rgb}{0.941, 0.243, 0.243}
\definecolor{Green7}{RGB}{55, 178, 77}
\definecolor{CubsRed}{RGB}{198, 1, 31}
\definecolor{BrickRed}{rgb}{0.6,0,0}
\definecolor{RoyalBlue}{rgb}{0,0,0.8}
\definecolor{BostonGreen}{RGB}{35, 138, 68}
\definecolor{CYRed}{RGB}{228, 0, 43}
\definecolor{CYPurple}{RGB}{215, 153, 93}
\definecolor{plgray}{HTML}{999999}

\definecolor{baselinecolor}{HTML}{EEEEEE}
\newcommand{\baseline}[1]{\cellcolor{baselinecolor}{#1}}

\newcommand{\cmark}{\textcolor{Green7}{\ding{51}}}%
\newcommand{\xmark}{\textcolor{red}{\ding{55}}}%

\newcommand{\pl}[1]{{\color{plgray} #1}}

\title{DEAS: DEtached value learning with \\ Action Sequence for Scalable Offline RL}
\newcommand{\metfull}{DEtached value learning with Action Sequence\xspace} 
\newcommand{\metabbr}{DEAS\xspace} 
\newcommand{\placeholder}[1]{{\color{lightgray}\lipsum[1]}}

\vspace{-0.1in}
\author{Changyeon Kim$^1$\thanks{Work done while visiting University of Texas at Austin.\ \hfill \textbf{Project page:} {\scriptsize\url{https://changyeon.site/deas}}}{\quad}Haeone Lee$^1${\quad}Younggyo Seo$^2${\quad}Kimin Lee$^{1}$\thanks{Equal advising.}{\quad}Yuke Zhu$^{3,4}$$^\dagger$\\
$^1$KAIST \,\,$^2$UC Berkeley\,\,\,$^3$University of Texas at Austin\,\,\,$^4$NVIDIA\,\,\, \\
}

\iclrfinalcopy 
\begin{document}

\maketitle

\vspace{-0.2in}
\begin{abstract}
    Offline reinforcement learning (RL) presents an attractive paradigm for training intelligent agents without expensive online interactions. However, current approaches still struggle with complex, long-horizon sequential decision making. In this work, we introduce \textit{\metfull} (\metabbr), a simple yet effective offline RL framework that leverages action sequences for value learning. These temporally extended actions provide richer information than single-step actions and can be interpreted through the options framework via semi-Markov decision process Q-learning, enabling reduction of the effective planning horizon by considering longer sequences at once. However, directly adopting such sequences in actor-critic algorithms introduces excessive value overestimation, which we address through detached value learning that steers value estimates toward in-distribution actions that achieve high return in the offline dataset. We demonstrate that \metabbr consistently outperforms baselines on complex, long-horizon tasks from OGBench and can be applied to enhance the performance of large-scale Vision-Language-Action models that predict action sequences, significantly boosting performance in both RoboCasa Kitchen simulation tasks and real-world manipulation tasks.

\end{abstract}
\vspace{-0.1in}

\section{Introduction}
\vspace{-0.075in}
    Offline reinforcement learning (RL)~\citep{lange2012batch, levine2020offline} enables learning from static datasets without incurring online data collection risks, while circumventing the need for expensive expert demonstrations. However, existing methods primarily focus on short-horizon tasks with dense rewards~\citep{yu2020meta, fu2020d4rl, gulcehre2020rl, mandlekar2021matters} and struggle to scale to complex long-horizon scenarios. Recent attempts using large-scale architectures~\citep{kumar2023offline, kumar2022pre, chebotar2023q, springenberg2024offline} show promise, but their effectiveness on complex tasks remains unexplored.
    
    To address the need for long-horizon evaluation, recent work~\citep{park2025ogbench, park2025horizon} has proposed challenging benchmarks for complex offline RL and demonstrated that reducing the effective planning horizon (i.e., shortening the time span over which the agent must plan) in both value and policy learning via $n$-step TD updates with high $n$ values and hierarchical policies is essential. However, these approaches rely on goal-conditioned RL with explicit expert-provided goals, which are often unavailable in practice. For instance, high $n$ values in $n$-step TD updates introduce increased bias and bootstrap error in standard RL without explicit goal information~\citep{tsitsiklis1996analysis, kearns2000bias, sutton2018reinforcement}. 
    
    These limitations underscore the need for alternative approaches to horizon reduction (reducing the planning horizon) 
    that work without explicit goal conditioning. One promising direction is leveraging action sequences, which have shown 
    success in behavior cloning~\citep{pomerleau1988alvinn} for capturing noisy, temporally-relevant distributions in 
    expert demonstrations~\citep{chi2023diffusion, zhao2023learning}. However, existing attempts to use action sequences for RL remain insufficient for achieving robust horizon reduction. Q-chunking~\citep{li2025reinforcement} has explored the use of action sequences for RL, demonstrating their potential for temporally consistent exploration. However, introducing action sequences to standard actor-critic frameworks causes severe value overestimation~\citep{seo2024coarse} due to actors maximizing over potentially erroneous critic estimates with widely spanned action spaces. This problem is exacerbated in offline RL where distribution shift creates extrapolation errors~\citep{kumar2019stabilizing,fujimoto2019off,kumar2020conservative}. While CQN-AS~\citep{seo2024coarse} proposes a value-only approach to avoid this issue, it introduces discretization errors that limit performance in complex tasks and cannot leverage expressive policy classes~\citep{wang2023diffusion,hansen2023idql,park2025flow}. For this reason, our research aims to develop methods that can leverage action sequences for horizon reduction while avoiding value overestimation and maintaining compatibility with expressive policy architectures.

    \begin{figure*}[t]
    \centering
    \includegraphics[width=0.87\textwidth]{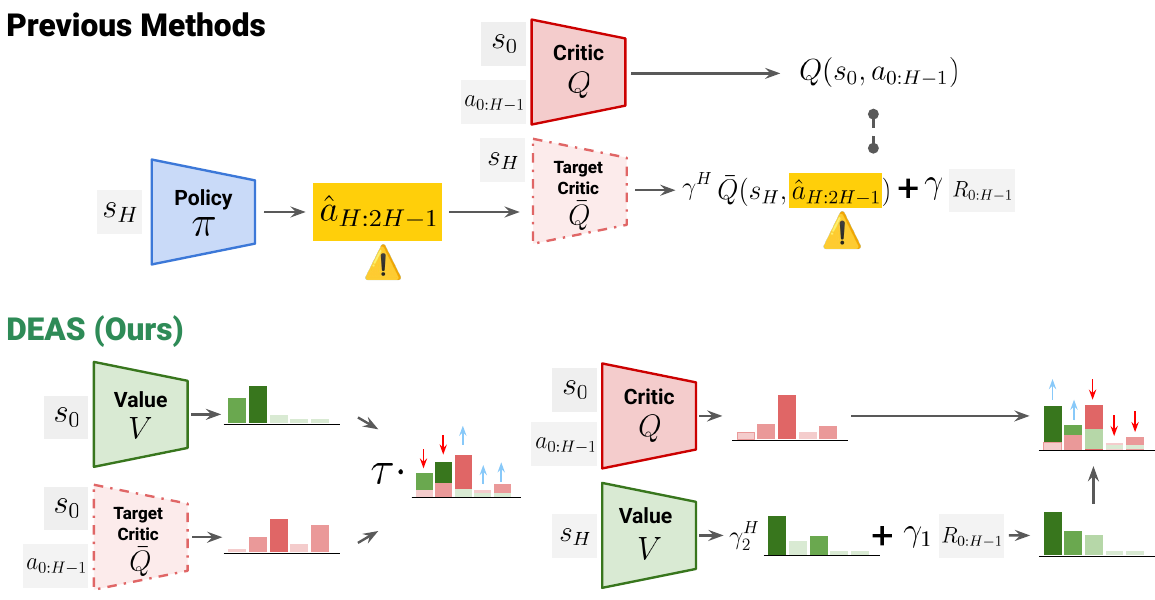}
    \caption{\textbf{Overview.} \metabbr is an offline RL framework that learns from action sequences instead of single actions. Unlike previous methods that couple actor-critic training, our key insight is to train the critic separately from the policy (detached value learning) using action sequences, which enables stable learning while avoiding value overestimation. We further enhance stability by combining distributional RL objectives and using dual discount factors, which leads to additional improvement.} 
    \label{figure:method}
    \vspace{-0.2in}
\end{figure*}
    \vspace{-0.1in}
    \paragraph{Our approach} We present \textit{\metfull} (\metabbr), an offline RL framework that leverages action sequences for scalable value learning in complex tasks. Our method treats consecutive action timesteps as inputs to the value function, implementing the simplest form of the options framework~\citep{sutton1999between,stolle2002learning}. This design provides principled horizon reduction analogous to $n$-step TD updates with temporally extended actions, while action sequences offer richer information than single-step actions without requiring explicit goal conditioning. To address the value overestimation challenges inherent in learning value functions with action sequences in offline RL settings, we employ detached value learning~\citep{kostrikov2022offline} that decouples critic training from the actor, biasing value estimates toward high-return actions present in the offline dataset. This method is appealing as it can be applied to any expressive policy architectures including large-scale Vision-Language-Action models (VLAs) without the hazard of value overestimation. Additionally, we propose to incorporate distributional RL~\citep{farebrother2024stop} in value learning to mitigate instability from accumulated bias in multi-step returns.
    
    We validate \metabbr through comprehensive experiments on challenging long-horizon tasks from OGBench~\citep{park2025ogbench}, where standard offline RL methods struggle to achieve meaningful success rates. Our method consistently outperforms all baselines, demonstrating its effectiveness on complex tasks. Additionally, we show that \metabbr can be used to improve the performance of VLAs~\citep{bjorck2025gr00t} in hard tasks from RoboCasa Kitchen~\citep{nasiriany2024robocasa} and real-world manipulation tasks, which significantly improves performance compared to policies trained solely on expert demonstrations. These results demonstrate \metabbr's practical applicability and potential for scaling offline RL to real-world scenarios.

    \paragraph{Contributions} We highlight the key contributions of our paper below:
    \vspace{-0.1in}
    \begin{itemize}[leftmargin=4mm]
        \item We present \metabbr: \textit{\metfull}, a simple yet effective offline RL method that leverages action sequences for training critics and employs detached value learning with classification loss for stable training.
        \item We demonstrate that \metabbr significantly outperforms baselines on complex, long-horizon tasks across 30 diverse scenarios in OGBench~\citep{park2025ogbench}.
        \item We demonstrate that \metabbr can enhance the performance of large-scale VLAs, achieving superior results on complex tasks from RoboCasa Kitchen~\citep{nasiriany2024robocasa} and real-world manipulation tasks compared to policies trained solely on expert demonstrations.
    \end{itemize}


\section{Related Work}
\vspace{-0.075in}
    \paragraph{Offline reinforcement learning}
    Offline RL focuses on learning policies from fixed datasets without further environment interaction~\citep{levine2020offline}. The primary challenge lies in the distributional shift between the behavior policy and the learned policy, which can result in value overestimation and suboptimal performance. Previous work has proposed various approaches including weighted regression~\citep{peng2019advantage, nair2020awac, wang2020critic}, conservative regularization~\citep{kumar2020conservative}, behavioral regularization~\citep{fujimoto2019off,fujimoto2021minimalist,tarasov2023revisiting,park2025flow}, and in-sample distribution maximization~\citep{kostrikov2022offline,xu2023offline,garg2023extreme}. Our method builds upon in-sample distribution maximization approaches, particularly IQL~\citep{kostrikov2022offline}, extending them to handle action sequences while maintaining stability by removing the critic update based on the actor's output. Furthermore, our method has the advantage of being adaptable to any policy extraction method for the final policy, making it more flexible and practical.

\vspace{-0.075in}
    \paragraph{BC/RL with action sequence}
    Adopting action sequence has been actively investigated in both imitation learning and RL. Behavior cloning advances show that predicting action sequences captures temporal dependencies from expert demonstrations that single-step actions miss~\citep{chi2023diffusion,zhao2023learning,black2025pi_0,bjorck2025gr00t,intelligence2025pi05visionlanguageactionmodelopenworld}. Several works have introduced action sequences into RL~\citep{li2024top,tian2025chunking}, with Q-Chunking~\citep{li2025reinforcement} demonstrating incorporation into actor-critic frameworks in offline-to-online RL without policy class constraints. However, this approach faces fundamental challenges: expanded action spaces increase value overestimation risk, particularly in offline settings with limited data coverage~\citep{kumar2019stabilizing}, yet this issue remains unaddressed. CQN-AS~\citep{seo2024coarse} circumvents this by removing the actor entirely, but introduces accumulating discretization errors that severely limit performance in complex tasks and prevent use of expressive policy classes~\citep{wang2023diffusion,park2025flow}. Our approach uniquely combines both paradigms: we leverage horizon reduction from action sequences while addressing value overestimation through detached value learning, enabling stable training with any policy architecture.

\vspace{-0.075in}
\section{Preliminaries}
\vspace{-0.075in}
    \paragraph{Problem formulation} We consider a Markov Decision Process (MDP)~\citep{sutton2018reinforcement} $\mathcal{M} = (\mathcal{S}, \mathcal{A}, p, R, \rho_0, \gamma)$, where $\mathcal{S}$ is the state space, $\mathcal{A}$ is the action space, $R(s,a): \mathcal{S} \times \mathcal{A} \rightarrow \mathbb{R} $ is the reward function, $p(s'|s, a): \mathcal{S} \times \mathcal{A} \rightarrow \Delta (\mathcal{S})$ is the transition function, $\rho_0$ is the initial state distribution, and $\gamma$ is the discount factor. In this paper, we focus on offline reinforcement learning, where we have access only to a static dataset $\mathcal{D} = {\{ \tau^{i} \}}_{i=0}^{N}$ containing $N$ trajectories of fixed length $H$, where each trajectory $\tau^{i} = (s_0, a_0, r_0, \ldots, s_H, a_H, r_H)$ represents a sequence of states, actions, and rewards. The dataset is collected using a data collection policy $\pi_{\mathcal{D}}: \mathcal{S} \rightarrow \Delta(\mathcal{A})$, which may be unknown or suboptimal. Unlike online RL, we cannot interact with the environment during training. The objective is to learn a policy $\pi: \mathcal{S} \rightarrow \Delta(\mathcal{A})$ that maximizes the expected sum of discounted rewards $\mathbb{E}_{\rho_0, \pi, p}\left[\sum_{t=0}^{\infty} \gamma^{t}R(s_t, a_t)\right]$ using only this fixed dataset. 

\vspace{-0.07in}
    \paragraph{Options framework} To formalize the idea for flexible temporal abstractions in RL and MDP, a Markovian option $\omega \in \Omega$ is defined as a triplet $(\mathcal{I}_{\omega}, \pi_{\omega}, \beta_{\omega})$. $\mathcal{I}_{\omega} \subseteq \mathcal{S}$ is the initiation set, $\pi_{\omega}$ is an \textit{intra-option} policy, and $\beta_{\omega} : \mathcal{S} \rightarrow [0, 1]$ is the termination function. 
    For any MDP $\mathcal{M}$ and any Markovian option $\omega_{\mathcal{M}}$ defined on $\mathcal{M}$, a decision process that follows only the option can be configured as an Semi-Markovian Decision Process (SMDP), which guarantees the existence of a set of optimal policies, denoted as $\Pi^*_{\omega}$. For more detailed explanations and proofs, please refer to \citet{sutton1999between}.

\vspace{-0.075in}
    \paragraph{Implicit Q Learning (IQL)~\citep{kostrikov2022offline}} Instead of regularizing the critic with the actor output, IQL approximates the optimal critic to be maximized only in the region of action distributions present in the offline dataset with an in-sample expectile regression. Given a parameterized critic $Q(s_t, a_t; \theta)$, target critic $Q(s_t, a_t; \bar{\theta})$, and value network $V(s_t; \psi)$, the objective for value learning is defined as:
    \begin{gather*}
        \mathcal{L}_{V}(\psi) = \mathbb{E}_{(s_t, a_t) \sim \mathcal{D}} \left[L^{\tau}_2(\bar{Q}(s_t, a_t; \bar{\theta})- V(s_t; \psi))\right] \label {eq:value_loss_regress_single} \\
        \mathcal{L}_{Q}(\theta) = \mathbb{E}_{(s_t, a_t) \sim \mathcal{D}} \left[(R(s,a) + \gamma V(s_{t+1}; \psi) - Q(s_t, a_t; \theta))^2\right] \label{eq:critic_loss_regress_single}
    \end{gather*}
    where $L^{\tau}_2(u) = |\tau - \mathbbm{1}(u < 0)|u^2$ is the expectile loss with expectile parameter $\tau \in [0, 1]$. By using $\tau > 0.5$, Equation~\ref{eq:value_loss_regress_single} penalizes the overestimated value in out-of-distribution actions, letting $V$ and $Q$ to be only approximated in the region of in-distribution actions.

\section{Method}

    We propose \metfull (\metabbr), an offline RL method that models action sequences for scalable learning. Our approach consists of two key components: (1) a critic function $Q(s_t, o_t; \theta)$ that estimates expected returns for the option (consisting of $H$-step action sequence) $o_t := a_{t:t+H-1}$ from state $s_t$ under the data collection policy $\pi_{\mathcal{D}}$, and (2) a flexible policy update mechanism applicable to any policy $\pi(a_{t:t+H-1} ; s_t, \phi)$ that outputs $H$-step action sequences. Section~\ref{sec:options_framework} describes how we integrate action sequences into SMDP Q-learning for horizon reduction, while Section~\ref{sec:deas} introduces how \metabbr enables stable training through detached value learning, distributional RL, and dual discount factors. We provide pseudocode in Algorithm~\ref{alg:metabbr} and implementation details in Appendix~\ref{sec:implementation}.
    \subsection{Options framework for action sequence RL}
    \label{sec:options_framework}
        Complex tasks require coordinated action sequences where each action's effectiveness depends on its context within the sequence. For instance, in OGBench $\mathtt{puzzle}$ or $\mathtt{cube}$ tasks, success depends on planning through multiple intermediate steps and maintaining consistent actions over extended periods. These temporal dependencies and hidden sub-tasks are not captured by current state representations, making it challenging for agents to learn effective policies. The challenge becomes even greater in goal-free settings, where agents must discover these sequential patterns from offline data without explicit goal instructions.

        To address these challenges, we propose modeling consecutive action sequences as single decision units within the options framework. We treat each $H$-step action sequence $o_t:= a_{t:t+H-1} = \{a_t, a_{t+1}, \ldots, a_{t+H-1}\}$ as an option, which naturally induces a SMDP~\citep{bradtke1994reinforcement,feinberg1994constrained,sutton1999between,baykal2010semi} that guarantees the existence of an optimal policy. Specifically, the option $\omega^*$ is defined as:
        \begin{gather*}
            \omega^* = (\mathcal{I}_{\omega^*}, \pi_{\omega^*}, \beta_{\omega^*}) = (\mathcal{S}, \pi(o_t \mid s_t), \beta^*(s_t, k)) \\
            \beta^*(s_t, k) = 
            \begin{cases}
                1 & \text{if } k = H \\
                0 & \text{otherwise}
            \end{cases}
        \end{gather*}
        where $k$ denotes the number of steps executed within the current option. This leads to a Q-learning update rule that extends standard Q-learning~\citep{bradtke1994reinforcement}:
        \begin{gather*}
            Q(s_t, o_t;\theta) \leftarrow \sum_{k=0}^{H-1} \gamma_1^{k} R(s_t, a_{t+k}) + \gamma_2^{H} \max_{o' \in \mathcal{O}} Q(s_{t+H}, o';\theta)
            \label{eq:semi_q_learning}
        \end{gather*}
        where $\gamma_1$ and $\gamma_2$ are discount factors for intra-option and inter-option transitions, respectively. This formulation aggregates rewards over $H$ steps and propagates value estimates across temporally extended transitions, achieving horizon reduction similar to $n$-step TD learning~\citep{park2025horizon}. 

        \begin{algorithm}[t!]
            \caption{\metabbr}
            \label{alg:metabbr}
            \begin{algorithmic}
            \footnotesize
            
            \State \textbf{Required}: Offline dataset $\mathcal{D}$, Support range for return $\mathbf{v}_{\min}$, $\mathbf{v}_{\max}$, number of bins $m$, discount factor $\gamma_1, \gamma_2$
            \State Initialize parameters $\psi, \theta, \bar{\theta}, \phi$
            \While{not converged}
            \State Sample batch $\{(s_t, a_{t:t+H-1}, R_{t:t+H-1}, s_{t+H})\}$ from $\mathcal{D}$
            \State Compute the discounted return of intra option as $\hat{R}_{t:t+H} = \sum^{H-1}_{k=0}\gamma_1^{k}R(s_t, a_{t+k})$
            \State Compute $\bar{Q}(s,o;\bar{\theta})$ and $V(s;\psi)$ using equation (\ref{eq:distributional_critic})
            \BeginBox[fill=white]
            \LComment{Update V Network}
            \State Update $V(s;\psi)$ to minimize Equation~(\ref{eq:value_loss}) with $\bar{Q}(s,o;\bar{\theta})$ and $V(s;\psi)$
            \EndBox
            \BeginBox[fill=white]
            \LComment{Update Q Network}
            \State Update $Q(s,o;\theta)$ to minimize Equation~(\ref{eq:critic_loss})
            \EndBox
            \BeginBox[fill=white]
            \LComment{Update Actor Network}
            \State Update $\pi(s;\phi)$ with any type of policy extraction algorithms (e.g., BoN, DPG, AWR, etc.)
            \EndBox
            \State Update $\bar{\theta} = (1 - \beta) \cdot \bar{\theta} + \beta \cdot \theta$
            \EndWhile
            \Return $\pi(\pl{s})$
        
            \end{algorithmic}
        \end{algorithm}
        \vspace{-0.05in}

\subsection{\metabbr: \metfull}
\label{sec:deas}

    \paragraph{Detached value learning for handling action sequence} Action sequences introduce challenges for value function approximation, as the expanded action space makes it harder for the critic to estimate Q-values accurately. Meanwhile, the actor can exploit regions where the critic makes prediction errors, leading to value overestimation and unstable learning~\citep{seo2024coarse}. To address this, we adopt detached value learning~\citep{kostrikov2022offline,xu2023offline,garg2023extreme} that decouples actor and critic training, introducing a critic $Q(s_t, o_t ; \theta)$ and a value $V(s_t ; \psi)$ networks with the following losses following IQL~\citep{kostrikov2022offline}:
    \begin{gather*}
        \mathcal{L}_{V}(\psi) = \mathbb{E}_{(s_t, o_t) \sim \mathcal{D}} \left[L^{\tau}_2(\bar{Q}(s_t, o_t ; \bar{\theta}) - V(s_t; \psi))\right] \label{eq:value_loss_as} \\
        \mathcal{L}_{Q}(\theta) = \mathbb{E}_{(s_t, o_t) \sim \mathcal{D}} \left[(\hat{R}_{t:t+H-1} + \gamma_2^{H} V(s_{t+H}; \psi) - Q(s_t, o_t ; \theta))^2 \right] \label{eq:critic_loss_as},
    \end{gather*}
    where $\hat{R}_{t:t+H-1} = \sum_{k=0}^{H-1}{\gamma_1}^{k}R(s_t, a_{t+k})$ is the discounted return for the action sequence. This approach steers the critic toward high-return actions in the offline dataset without the potential of exploiting critic approximation errors, preventing value overestimation and enabling stable value learning even with longer action sequences.

    \paragraph{Distributional RL for enhanced stability} Even with detached value learning, the cumulative reward term $\hat{R}_{t:t+H-1}$ could introduce significant variance when $H$ is large. To enhance stability, we extend our framework with distributional RL~\citep{bellemare2017distributional,farebrother2024stop}, modeling both critic and value networks as categorical distributions over fixed support $[\mathbf{v}_{\min}, \mathbf{v}_{\max}]$ discretized into $m$ bins:
    \begin{equation}
        Q(s, o ; \theta) = \mathbb{E}\left[ Z(s,o ; \theta)\right] \quad Z(s,o;\theta) = \sum_{i=1}^{m}\hat{p}_{i}(s,o;\theta) \cdot \delta_{z_i} \quad \hat{p}_{i}(s,o;\theta) = \frac{e^{l_i(s,o;\theta)}}{\sum_{i=1}^{m}e^{l_i(s,o;\theta)}},
        \label{eq:distributional_critic}
    \end{equation}
    where $V(s;\psi)$ is computed similarly, by conditioning only on the state $s$.
    To address scale differences between regression and classification objectives, we maintain IQL's weighting scheme but replace regression with classification-based learning:
    \begin{equation}
    \begin{split}
        \mathcal{L}_{V}(\psi) &=  \mathbb{E}_{(s_t, o_t) \sim \mathcal{D}} \left[ \alpha_{t} \cdot \sum^{m}_{i=1}\hat{p}_{i}(s_t ; \psi)\log{\hat{p}_{i}(s_t, o_t ; \bar{\theta})}\right] \\
        \alpha_{t} &= 
        \begin{cases}
            \tau & \text{if } \bar{Q}(s_t, o_t ; \bar{\theta}) \geq V(s_t ; \psi) \\
            1 - \tau & \text{otherwise},
        \end{cases}
    \end{split}
    \label{eq:value_loss}
    \end{equation}
    \vspace{-0.05in}
    \begin{align}
        \mathcal{L}_{Q}(\theta) = \mathbb{E}_{(s_t, a_{t:t+H-1},s_{t+H}) \sim \mathcal{D}} \left[ \sum^{m}_{i=1}p_{i}(s_t ; \psi)\log{\hat{p}_{i}(s_t, a_{t:t+H-1} ; \hat{\theta})}\right]. \label{eq:critic_loss}
    \end{align}
    For target probabilities $p_i$, we adopt the truncated normal distribution with mean as Bellman target $(\mathcal{\hat{T}V})(s,a_{t:t+H-1}) = \sum_{k=0}^{H-1} \gamma_1^{k} r_{t+k} + \gamma_2^{H}V(s_{t+H} ; \psi)$ and standard deviation $\sigma = 0.75 \cdot (v_{max} - v_{min} / m)$, inspired by~\citet{farebrother2024stop}.

    \paragraph{Dual discount factors} To further enhance stability and expressiveness in value estimation, we employ two separate discount factors: $\gamma_1$ for intra-option (within action sequence) rewards and $\gamma_2$ for inter-option (across action sequences) rewards. This dual-discounting scheme enables the value function to appropriately weigh immediate and future returns, mitigating issues such as value explosion or collapse that can arise from improper scaling of returns. In our experiments, we observe that decreasing the intra-option discount factor $\gamma_1$ and increasing the inter-option discount factor $\gamma_2$ leads to more stable training, and is critical for stable training, especially when the action sequence becomes longer (see Section~\ref{sec:ablation} for the supporting results).

    \paragraph{Compatible policy methods} For obtaining final policy $\pi(s;\phi)$, our framework is compatible with a variety of policy extraction strategies~\citep{park2024value}, including weighted behavior cloning~\citep{peng2019advantage}, deterministic policy gradient (DPG)~\citep{fujimoto2021minimalist}, best-of-N sampling~\citep{chen2023offline}, and flow-matching approaches~\citep{park2025flow}. Since value function training does not require querying the policy, it can be performed independently, and the policy can be updated separately. To demonstrate this, we illustrate the effectiveness of our method using various policy extraction methods in our experiments.

\begin{figure}
    \centering
    \begin{subfigure}[b]{0.495\textwidth}
        \centering
        \includegraphics[width=\textwidth]{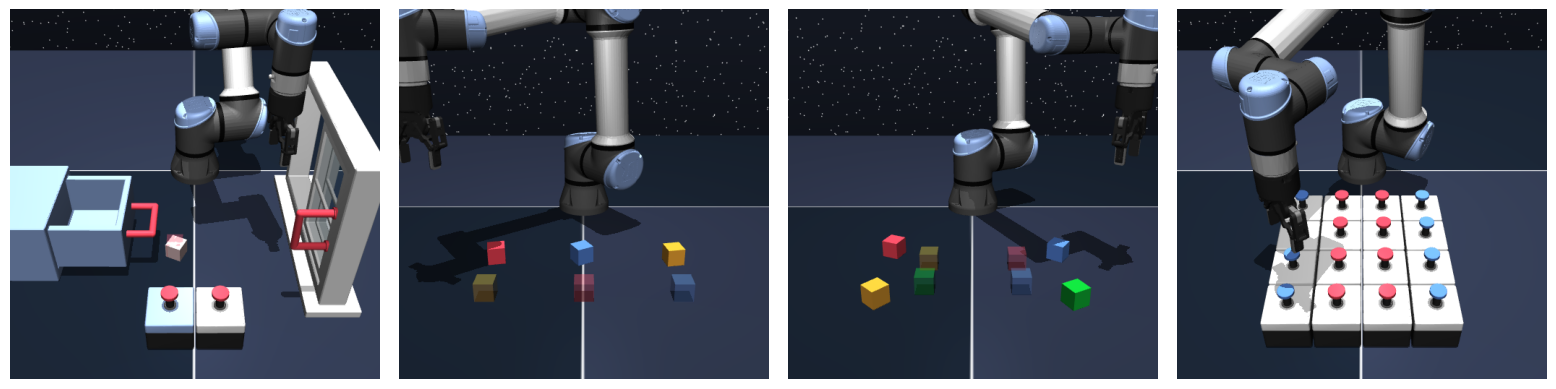}
        \label{fig:example_ogbench}
    \end{subfigure}
    \begin{subfigure}[b]{0.495\textwidth}
        \centering
        \includegraphics[width=\textwidth]{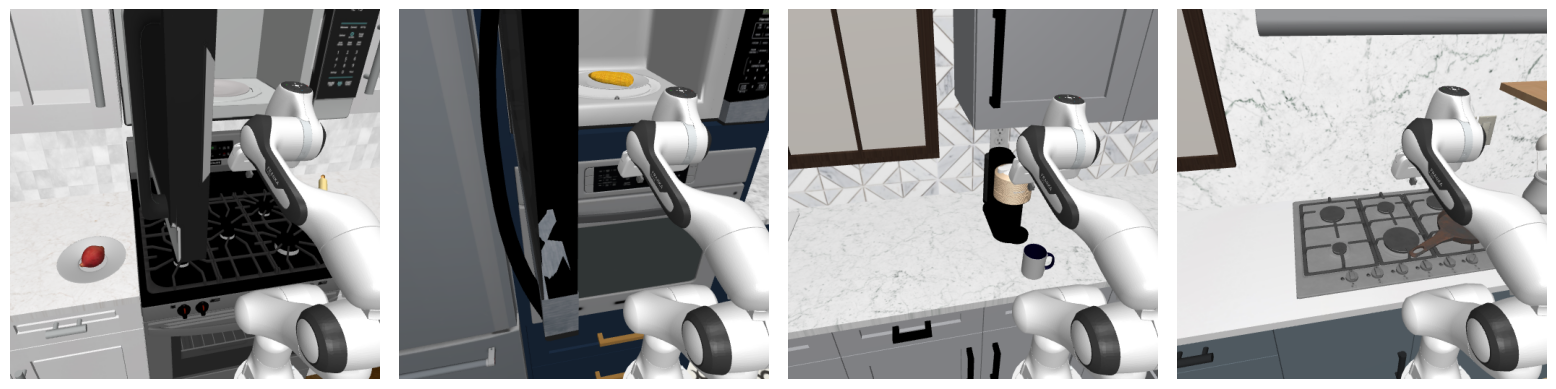}
        \label{fig:example_robocasa}
    \end{subfigure}
 
    \vspace{-0.15in}
    \caption{
        \textbf{Simulation task examples.} We study \metabbr on 30 different tasks from OGBench~\citep{park2025ogbench} and 4 challenging manipulation tasks from RoboCasa Kitchen~\citep{nasiriany2024robocasa}.} 
    \label{fig:example_tasks}
    \vspace{-0.15in}
\end{figure}

\section{Experiments}
\label{sec:exp}
    \vspace{-0.075in}
    We first validate the effectiveness of \metabbr through extensive experiments on various complex tasks in OGBench~\citep{park2025ogbench}. Additionally, to prove that \metabbr can be naturally plugged into large-scale VLAs for practical applications, we evaluate \metabbr by fine-tuning GR00T N1.5~\citep{gr00tn1_5} using offline RL methods on 4 hard tasks from RoboCasa Kitchen~\citep{nasiriany2024robocasa} and also conduct real-world experiments with Franka Emika Research 3 Robot Arm. See Figure~\ref{fig:example_tasks} and Figure~\ref{fig:fr3_example} for task examples used in our experiments.
    \vspace{-0.075in}
    \subsection{OGBench Experiments}
    \paragraph{Setup} We evaluate on 6 manipulation environments from OGBench~\citep{park2025ogbench}, each with 5 subtasks. We use datasets ranging from 1M to 100M transitions based on task difficulty. While OGBench is originally designed for offline goal-conditioned RL, we use its single-task variants ($\tt{`-singletask'}$) for reward-maximizing offline RL. For fair comparison, all methods use identical MLP architectures for actor networks and adopt the same policy extraction approach as FQL~\citep{park2025flow}, except for CQN-AS, which uses value function networks as the actor itself through discretization. Action sequence length $H$ is set to 8 for $\tt{scene}$ and $\tt{puzzle}$ tasks, and 4 for $\tt{cube}$ tasks, with $n=H$ is used for $n$-step FQL. More details about the experimental setup can be found in Appendix~\ref{sec:ogbench_implementation}.
    \vspace{-0.075in}
    \paragraph{Baselines} We compare against \textcolor{CubsRed}{FQL}~\citep{park2025flow}, a state-of-the-art offline RL method using one-step distillation between flow matching models with different denoising steps, and \textcolor{CubsRed}{$n$-step FQL}~\citep{sutton2018reinforcement}, which extends FQL with $n$-step TD updates for horizon reduction~\citep{park2025horizon}. While increasing $n$ increases bias in standard offline RL, \metabbr\ explicitly models action sequences while maintaining horizon reduction benefits. We also consider \textcolor{CubsRed}{Q-Chunking (QC)}~\citep{li2025reinforcement}, which uses action chunking for actor-critic training while keeping the interaction between actor and critic, while \metabbr\ uses detached value learning. For fair comparison with ours, we extensively tune QC-FQL hyperparameters to achieve higher performance than the original paper. Lastly, \textcolor{CubsRed}{CQN-AS}~\citep{seo2024coarse}, a value-based RL method with action sequence utilizing multi-level critics with iterative discretization, is included as a baseline.
    \vspace{-0.075in}
    \begin{table}[t]
    \caption{\textbf{Offline RL results} in 6 task categories from OGBench~\citep{park2025ogbench}. We report the success rate (\%) and 95\% stratified bootstrap confidence interval over 4 runs. \textbf{Bold} indicates the values at or above 95\% of the best performance. Please refer to Table~\ref{tab:full} for the full results. }
    \centering

    \resizebox{\textwidth}{!}{
    \begin{tabular}{lccccc>{\columncolor{green!10}}c}
        \toprule
         Task Category & \#Data & FQL & N-step FQL & QC-FQL & CQN-AS & \textbf{\metabbr} \\
         \midrule
         \multicolumn{1}{l}{\tt{scene-play-singletask ($\mathbf{5}$ tasks)}} & \multirow{3}{*}{1M} & $50$ {\tiny $\pm 3$}   & $36$ {\tiny $\pm 2$}  & $\mathbf{73}$ {\tiny $\pm 2$} & $1$ {\tiny $\pm 1$} & $\mathbf{76}$ {\tiny $\pm 2$}\\
         \multicolumn{1}{l}{\tt{cube-double-play-singletask ($\mathbf{5}$ tasks)}} &  & $14$ {\tiny $\pm 2$}   & $4$ {\tiny $\pm 2$}  & $41$ {\tiny $\pm 3$} & $2$ {\tiny $\pm 1$} & $\mathbf{48}$ {\tiny $\pm 2$}\\
         \multicolumn{1}{l}{\tt{puzzle-3x3-play-singletask ($\mathbf{5}$ tasks)}} &  & $44$ {\tiny $\pm 3$} & $36$ {\tiny $\pm 3$} & $62$ {\tiny $\pm 7$} & $0$ {\tiny $\pm 0$} & $\mathbf{91}$ {\tiny $\pm 3$}\\
         \midrule
         \multicolumn{1}{l}{\tt{cube-triple-play-singletask ($\mathbf{5}$ tasks)}} & \multirow{2}{*}{10M} & $10$ {\tiny $\pm 3$} & $23$ {\tiny $\pm 2$} & $\mathbf{83}$ {\tiny $\pm 4$} & $0$ {\tiny $\pm 0$} & $\mathbf{82}$ {\tiny $\pm 5$}\\
         \multicolumn{1}{l}{\tt{puzzle-4x4-play-singletask ($\mathbf{5}$ tasks)}} &  & $32$ {\tiny $\pm 4$} & $19$ {\tiny $\pm 5$} & $69$ {\tiny $\pm 8$} & $0$ {\tiny $\pm 0$} & $\mathbf{82}$ {\tiny $\pm 6$}\\
         \midrule
         \tt{cube-quadruple-play-singletask ($\mathbf{5}$ tasks)} & 100M & $17$ {\tiny $\pm 8$} & $36$ {\tiny $\pm 10$} & $45$ {\tiny $\pm 7$} & $0$ {\tiny $\pm 0$} & $\mathbf{64}$ {\tiny $\pm 8$}\\
         \bottomrule
    \end{tabular}
    }
    \vspace{-0.1in}
    \label{tab:main}
\end{table}
    \begin{figure}[t]
    \centering
    \begin{subfigure}[b]{0.32\textwidth}  
        \centering 
        \includegraphics[width=\textwidth]{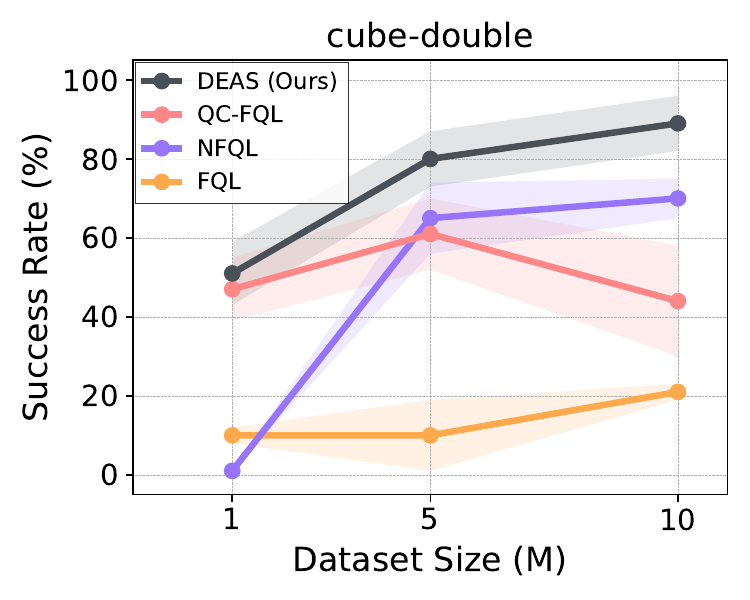}
        \label{fig:cube_double_scaling}
        \vspace{-0.175in}
    \end{subfigure}
    \begin{subfigure}[b]{0.32\textwidth}
        \centering
        \includegraphics[width=\textwidth]{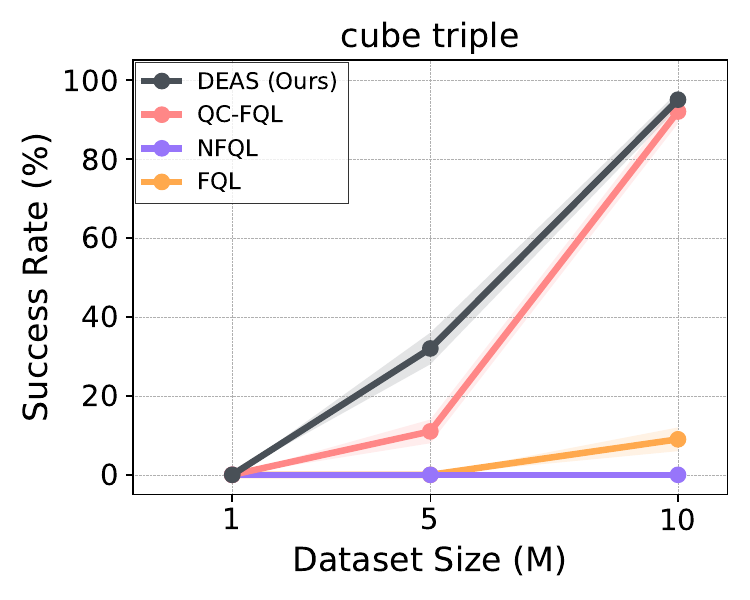}
        \label{fig:cube_triple_scaling}
        \vspace{-0.175in}
    \end{subfigure}
    \begin{subfigure}[b]{0.32\textwidth}   
        \centering 
        \includegraphics[width=\textwidth]{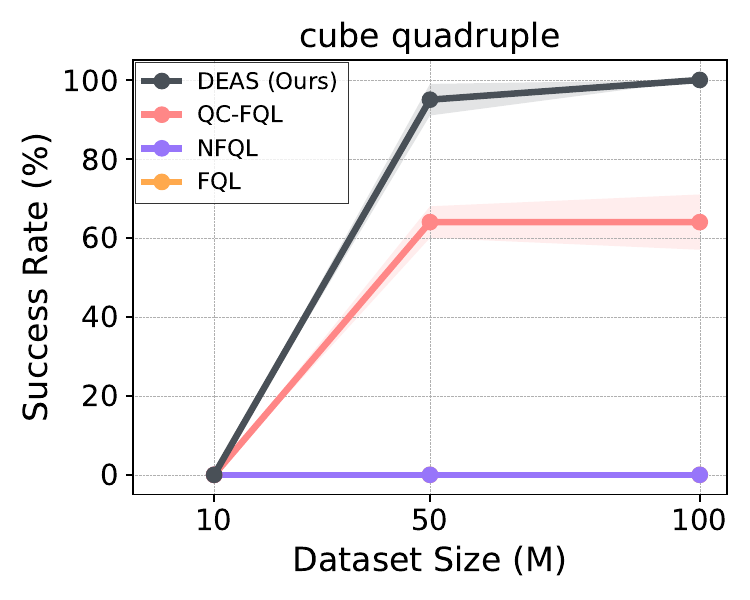}
        \label{fig:cube_quadruple_scaling}
        \vspace{-0.175in}
    \end{subfigure}
    \vspace{-0.1in}
    \caption{
    Agent performance across varying dataset sizes on three representative OGBench~\citep{park2025ogbench} tasks, evaluated by success rate (\%). Solid lines indicate the mean, while shaded areas denote the stratified bootstrap confidence intervals over 4 independent runs.
    }
    \label{fig:scaling}
    \vspace{-0.2in}
\end{figure}

    \paragraph{Quantitative results} As shown in Table~\ref{tab:main}, \metabbr consistently achieves the best performance across all 6 task categories with various dataset sizes. Comparing FQL and N-step FQL, we observe that simply increasing the $n$-step mostly leads to performance degradation due to bias in standard offline RL, while our detached value learning approach enables stable training with action sequences. Notably, \metabbr matches or outperforms QC-FQL across all tasks, demonstrating the effectiveness of our stable value learning in addressing offline RL instability. The method shows particularly strong performance on tasks requiring long-horizon reasoning like $\tt{puzzle}$ and the most challenging tasks (i.e., $\tt{cube-quadruple}$), where the benefits of using action sequences are most pronounced. CQN-AS shows significantly lower performance, likely due to its direct application of strong BC regularization on the value function in the presence of predominantly suboptimal data, along with cumulative errors from iterative discretizations that reduce action precision. 
    \vspace{-0.075in}

    \paragraph{Scaling analysis} To further validate the scalability of \metabbr, we conduct a scaling analysis on three representative OGBench tasks with varying dataset sizes. As shown in Figure~\ref{fig:scaling}, \metabbr consistently outperforms all baselines across all dataset sizes, achieving the highest success rates in every environment. The method demonstrates robust scaling across different dataset sizes, maintaining consistent performance gains even with larger datasets. This superior performance validates our approach of explicitly modeling action sequences while effectively leveraging suboptimal data through our detached value learning and stable multi-step training.
    \vspace{-0.075in}
\subsection{VLA Experiments}
\label{sec:vla}
\vspace{-0.075in}

    To validate the practical applicability of \metabbr, we demonstrate its effectiveness with large-scale VLAs~\citep{black2025pi_0,bjorck2025gr00t,gr00tn1_5}. These models, trained on internet-scale diverse datasets with billion-scale parameters, predict much longer action sequences and are widely used in robotics applications. However, deploying these models typically requires fine-tuning on task-specific data, which often necessitates collecting expensive expert demonstrations. We design our experiments to validate whether \metabbr can improve VLA performance by effectively utilizing suboptimal demonstrations alongside limited expert data, potentially reducing the required amount of costly expert demonstrations. See Appendix~\ref{sec:vla_exp} for more details.

    \subsubsection{RoboCasa Kitchen Experiments}
    \vspace{-0.05in}
    \paragraph{Setup} We employ GR00T N1.5~\citep{gr00tn1_5} as the backbone VLA. First, we fine-tune the VLA using 100 expert demonstrations from all 24 RoboCasa Kitchen tasks to verify that we achieve performance similar to the original GR00T N1~\citep{bjorck2025gr00t}. From these tasks, we select 4 tasks with the lowest success rates in their respective categories for our offline IL/RL experiments. We then collect 300 rollouts for each task from the resulting policy and apply various offline IL/RL methods. For RL methods, we fine-tune the base policy using behavior cloning on both expert demonstrations and the rollout dataset and use the model as an actor for training critic functions when necessary. For policy extraction, we adopt best-of-N sampling~\citep{chen2023offline,nakamoto2024steering}, where we sample multiple outputs from the policy and select the action sequence with the highest Q-value. We set $H=16$ for all methods, matching GR00T N1.5's action chunk size.
    \vspace{-0.075in}
    \begin{table}[t!]
    \centering\small
    \caption{
    \textbf{RoboCasa Kitchen evaluation results}. We fine-tune GR00T N1.5~\citep{gr00tn1_5} on 24 RoboCasa Kitchen tasks using 100 expert demonstrations per task. For 4 selected tasks, we collect 300 rollouts and apply offline IL/RL algorithms. Success rates (\%) on 50 episodes, aggregated with 3 seeds.  $\tt{PnPC2M}$ denotes `PnPCounterToMicrowave' and $\tt{PnPM2C}$ denotes `PnPMicrowaveToCounter'. \textbf{Bold} and \underline{underline} indicate best and runner-up results, respectively.}
    \vspace{-0.1in}
    \resizebox{0.9\textwidth}{!}{
    \begin{tabular}{l ccccc}
        \multicolumn{6}{r}{\small{$\dagger$ Reproduced performance}} \\
        \toprule
        Models & \tt{CoffeeSetupMug} & \tt{PnPC2M} & \tt{PnPM2C} & \tt{TurnOffStove} & Avg. \\
        \midrule
        \emph{Base models} \\
        GR00T N1$^{*}$
            & \phantom{0} $2.0$
            & \phantom{0} $0.0$ 
            & \phantom{0} $0.0$
            & \phantom{0} $15.7$
            & \phantom{0} $4.4$ \\
        GR00T N1.5${\dagger}$
            & \phantom{0} $4.7$
            & \phantom{0} $21.3$ 
            & \phantom{0} $7.3$
            & \phantom{0} $14.7$
            & \phantom{0} $12.0$ \\
        \arrayrulecolor{black!40}\midrule
        \emph{Imitation learning} \\
        \quad + Filtered BC
            & \phantom{0} $14.7$
            & \phantom{0} $25.3$
            & \phantom{0} $\underline{14.7}$
            & \phantom{0} $\textbf{19.3}$
            & \phantom{0} $18.5$ \\
        \arrayrulecolor{black!40}\midrule
        \emph{Offline RL} \\
        \quad + IQL
            & \phantom{0} $\underline{23.3}$
            & \phantom{0} $\underline{30.0}$
            & \phantom{0} $\underline{14.7}$
            & \phantom{0} $12.7$
            & \phantom{0} $\underline{20.2}$ \\
        \quad + QC
            & \phantom{0} $16.0$
            & \phantom{0} $28.7$
            & \phantom{0} $\underline{14.7}$
            & \phantom{0} $10.7$
            & \phantom{0} $17.5$ \\
        \rowcolor{green!10}\quad + \textbf{\metabbr (Ours)}
            & \phantom{0} $\mathbf{28.7}$
            & \phantom{0} $\mathbf{36.0}$
            & \phantom{0} $\mathbf{18.0}$
            & \phantom{0} $\underline{18.0}$
            & \phantom{0} $\mathbf{25.2}$ \\
        
        \bottomrule
    \end{tabular} 
    }
    \vspace{-0.15in}
    \label{tab:vla}
    \end{table}

    \paragraph{Baselines} We compare against several baselines across both imitation learning and reinforcement learning paradigms. For imitation learning, we consider \textcolor{CubsRed}{Filtered BC}, which fine-tunes the base policy using both expert demonstrations and successful episodes from the rollout data~\citep{oh2018self}. For reinforcement learning, we evaluate \textcolor{CubsRed}{IQL}, a value-based method that operates on single actions without requiring policy outputs. For determining action sequence in IQL, we use the very first action in the sequence for value estimation. Lastly, we consider \textcolor{CubsRed}{QC}, which employs action chunking for critic training but relies on predicted action sequences from VLA for the critic update. 
    \vspace{-0.075in}

    \paragraph{Results} As shown in Table~\ref{tab:vla}, \metabbr achieves the highest success rates in 3 out of 4 tasks, with the remaining task also showing improved performance compared to the base model. While filtered BC improves performance with simple approaches, but our approach exhibits additional performance gains by effectively utilizing suboptimal data. While single-step IQL also demonstrates effectiveness, it shows smaller performance gains across all tasks compared to our approach, due to its lack of understanding of action sequences. QC shows limited improvement compared to BC-based approaches, highlighting the advantage of our detached value learning with action sequences.
    \vspace{-0.075in}
    \subsubsection{Real-world Experiments} 
    \paragraph{Setup} We further investigate the effectiveness of \metabbr in real-world tasks using Franka Emika Research 3 Robot Arm. Inspired by RoboCasa Kitchen, we design pick-and-place tasks from the countertop to the bottom cabinet, with three different objects: $\tt{peach}$, $\tt{milka}$, and $\tt{hichew}$ (see Figure~\ref{fig:fr3_example}). For each task, we collect 5 demonstrations, fine-tune GR00T N1.5, collect 25 rollouts, and apply various offline IL/RL methods. We evaluate using 20 rollouts per task from 5 different initial points and use the same baselines as in the RoboCasa Kitchen experiments. Success rates are calculated based on partial success scoring (0-1 scale) that considers subtask completion, with detailed evaluation methodology provided in \Cref{sec:real_robot_exp}.
    \vspace{-0.075in}

    \paragraph{Results} In Table~\ref{tab:real_robot}, \metabbr achieves the highest success rates across all three pick-and-place tasks compared to baselines. The method shows consistent improvements, particularly on challenging objects like $\tt{milka}$ (a deformable object) where other approaches struggle. Notably, QC shows degraded performance compared to the base model, likely due to its instability when using action sequences with relatively small datasets, while our method shows stable improvement even with limited data. These results demonstrate that our detached value learning approach can be effectively applied to real-world robotic tasks and remains stable regardless of the dataset size. 
    \vspace{-0.075in}

\begin{table*}[t]
    \begin{minipage}{0.42\textwidth}
        \centering
        \small
        \captionof{figure}{\textbf{Real-world tasks.} We conduct pick-and-place tasks from the countertop to the bottom cabinet with $\tt{peach}$, $\tt{milka}$, and $\tt{hichew}$.}
        \includegraphics[width=0.32\textwidth]{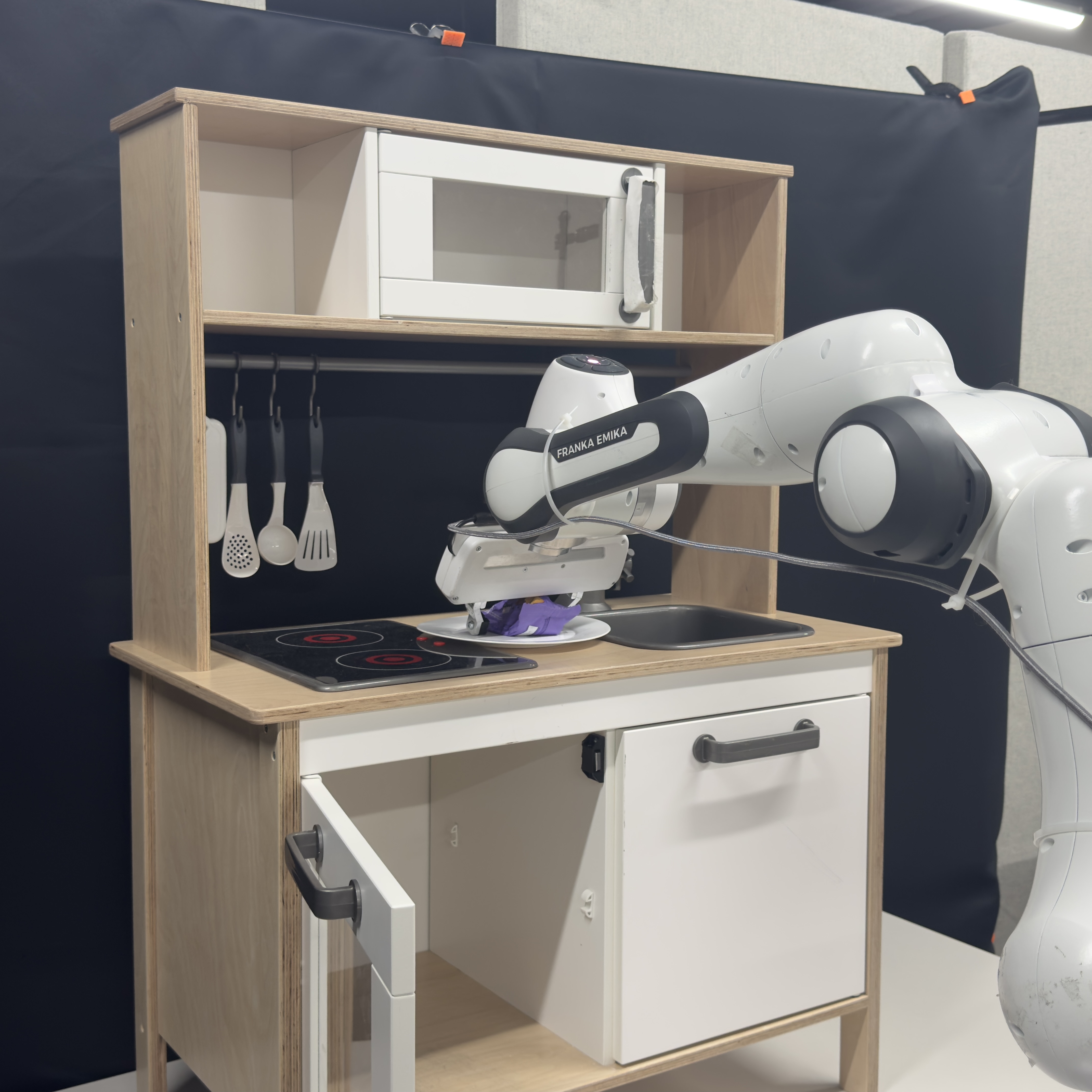}
        \includegraphics[width=0.32\textwidth]{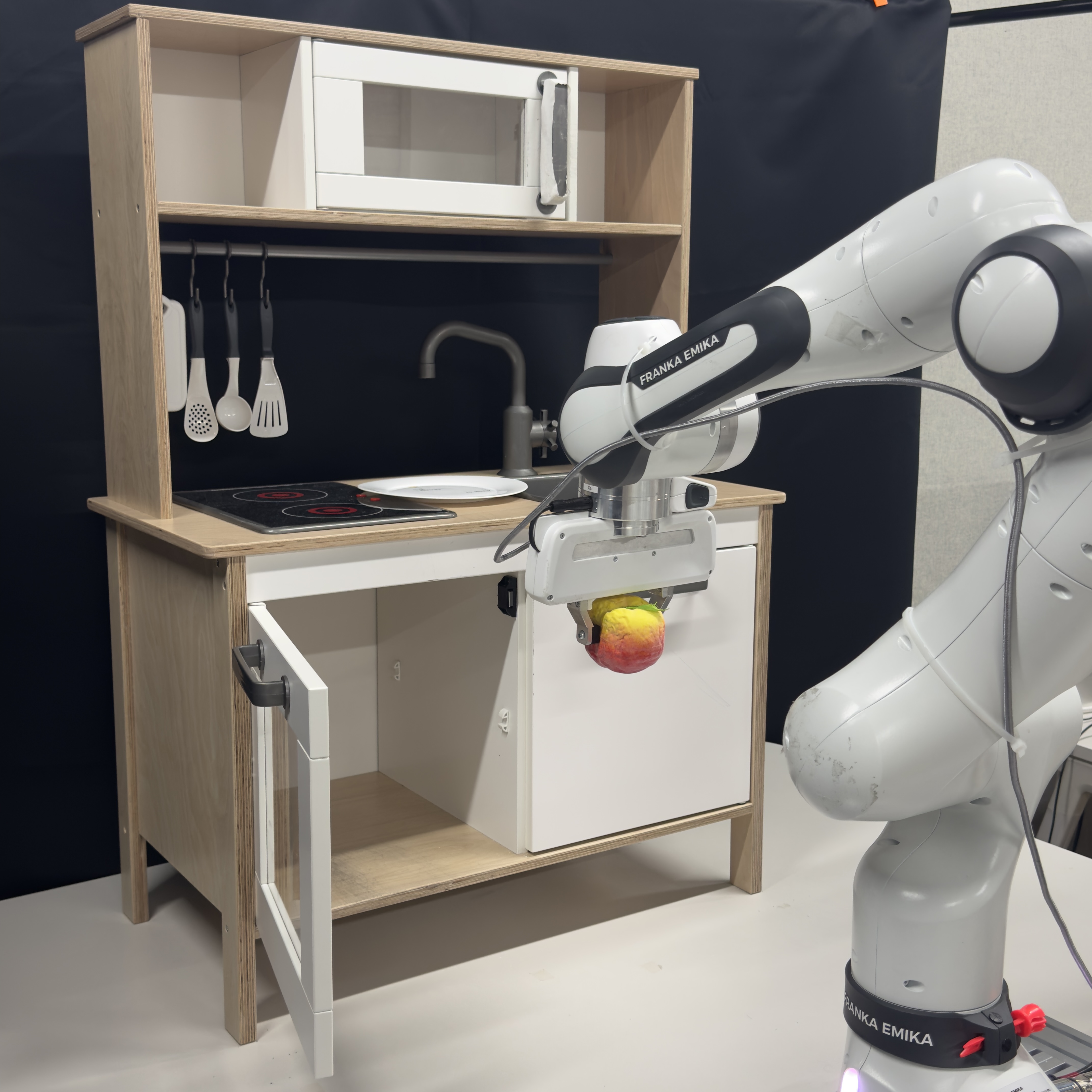}
        \includegraphics[width=0.32\textwidth]{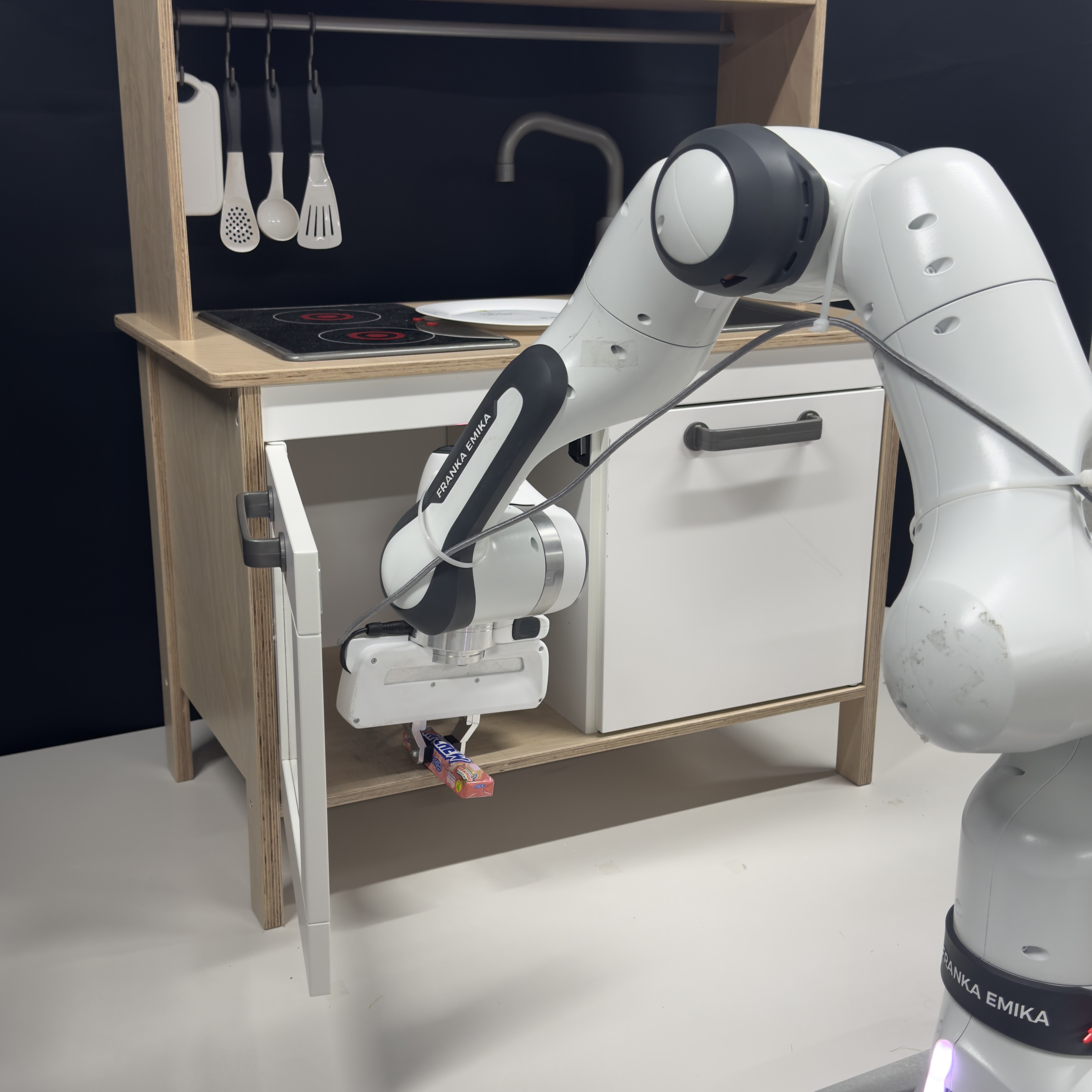}
        \\
        \centering
        \includegraphics[width=0.6\textwidth]{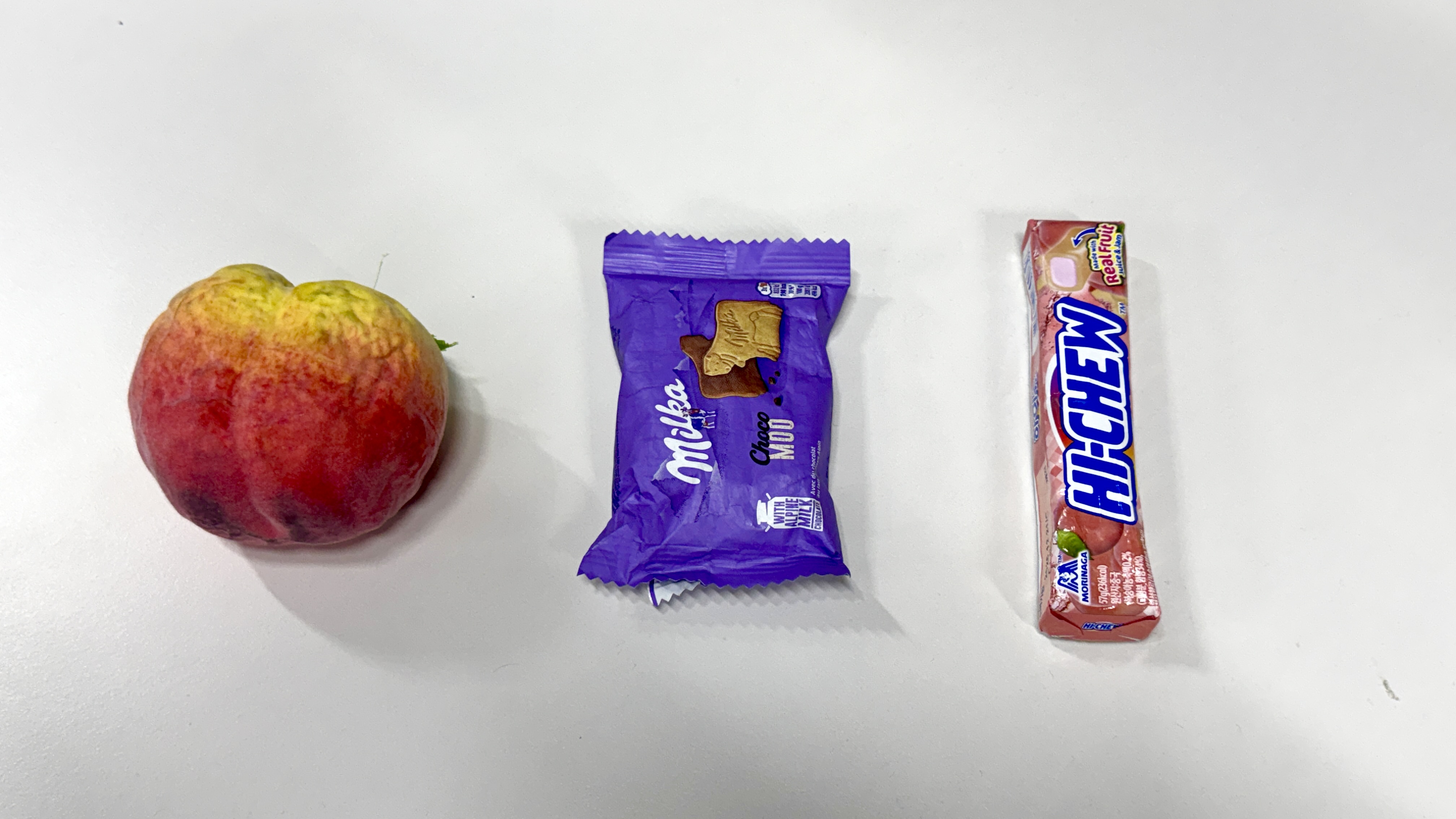}
        \label{fig:fr3_example}
    \end{minipage}
    \hspace{0.5em}
    \begin{minipage}{0.56\textwidth}
        \centering
        \small
        \captionof{table}{\textbf{Real-world evaluation results.}
            We report the partial success rate (\%, over 20 trials per task) on 3 tasks from 5 initial points.
            \textbf{Bold} and \underline{underline} indicate best and runner-up results, respectively.
        }\label{tab:real_robot}
        \resizebox{\textwidth}{!}{
        \begin{tabular}{lccccc}
            \toprule
        
            Models & \tt{peach} & \tt{milka} & \tt{hichew} & Avg. \\
            
            \midrule
            \emph{Base model} \\
            GR00T N1.5 & 62.0 & \underline{45.0} & 85.0 & 64.0 \\
            
            \arrayrulecolor{black!40}\midrule
            \emph{Imitation learning} \\
            +  Filtered BC & 76.3 & 25.0 & \underline{92.5} & 64.6 \\
            \arrayrulecolor{black!40}\midrule
            \emph{Offline RL} \\

            +  IQL & \underline{82.5} & 37.5 & 78.8 & \underline{66.3} \\
            +  QC & 58.8 & 15.0 & 45.0 & 39.6 \\
            \rowcolor{green!10}\textbf{+ \metabbr (Ours)} & \textbf{86.3} & \textbf{53.8} & \textbf{95.0} & \textbf{78.4} \\
            
            \bottomrule
        \end{tabular}
        }
    \end{minipage}
    \vspace{-0.05in}
\end{table*}
\begin{table*}[t]
    \centering
    \hspace{-3em}
    \begin{minipage}{\textwidth}
    \centering
    \begin{minipage}{0.22\linewidth}
        \centering
        \subfloat[
        Action sequence
        \label{table:effect_of_as}
        ]{
        \scriptsize
        \begin{tabular}{ccc}
        $H$ & Actor & SR \\ 
        \shline
        1 & 512 $\times$ 4 & $21$ {\tiny $\pm 3$} \\
        2 & 512 $\times$ 4 & $25$ {\tiny $\pm 5$} \\
        4 & 512 $\times$ 4 & $75$ {\tiny $\pm 8$} \\
        \baseline{8} & \baseline{512 $\times$ 4} & \baseline{$\mathbf{88}$ {\tiny $\pm 4$}} \\
        16 & 512 $\times$ 4 & $51$ {\tiny $\pm 4$} \\
        16 & 1024 $\times$ 4 & $84$ {\tiny $\pm 4$} \\
        \end{tabular}
        }
    \end{minipage}
    \hspace{0.5em}
    \begin{minipage}{0.22\linewidth}
        \centering
        \subfloat[
        Critic size
        \label{table:effect_of_critic_size}
        ]{
        \scriptsize
        \begin{tabular}{ccc}
        Critic & Value & SR \\ 
        \shline
        256 $\times$ 4 & 256 $\times$ 4 & $69$ {\tiny $\pm 7$} \\
        \baseline{512 $\times$ 4} & \baseline{256 $\times$ 4} & \baseline{$\mathbf{88}$ {\tiny $\pm 4$}} \\
        1024 $\times$ 4 & 256 $\times$ 4 & $\mathbf{91}$ {\tiny $\pm 4$} \\
        512 $\times$ 4 & 512 $\times$ 4 & $50$ {\tiny $\pm 4$} \\
        \end{tabular}
        \vspace{0.1in}
        }
    \end{minipage}
    \hspace{2em}
    \begin{minipage}{0.17\linewidth}
        \centering
        \subfloat[
        Objectives
        \label{table:objectives}
        ]{
        \scriptsize
        \begin{tabular}{ccc}
        IQL & HLG & SR \\ 
        \shline
        \xmark & \cmark & $75$ {\tiny $\pm 5$} \\
        \cmark & \xmark & $63$ {\tiny $\pm 6$}\\
        \baseline{\cmark} & \baseline{\cmark} & \baseline{$\mathbf{88}$ {\tiny $\pm 4$}} \\
        \end{tabular}
        \vspace{0.2in}
        }
    \end{minipage}
    \hspace{2em}
    \begin{minipage}{0.17\linewidth}
        \centering
        \subfloat[
        $\gamma_1$ and $\gamma_2$
        \label{table:discount_factor}
        ]{
        \scriptsize
        \begin{tabular}{ccc}
        \makecell{$\gamma_1$} & \makecell{$\gamma_2$} & SR \\ 
        \shline
        0.8 & 0.999 & $\mathbf{87}$ {\tiny $\pm 4$}\\
        \baseline{0.9} & \baseline{0.999} & \baseline{$\mathbf{88}$ {\tiny $\pm 4$}} \\
        0.99 & 0.999 & $81$ {\tiny $\pm 5$}\\
        0.999 & 0.999 & $80$ {\tiny $\pm 8$}\\
        \end{tabular}
        \vspace{0.1in}
        }
    \end{minipage}
    \end{minipage}
    \centering
    \caption{\textbf{Ablation studies.} We investigate the effect of (a) action sequence length $H$, (b) critic and value model size, (c) training objectives, and (d) separate discount factors $\gamma_1$ and $\gamma_2$ for intra-option and inter-option rewards. SR denotes success rate (\%) and default settings are highlighted in \colorbox{baselinecolor}{gray}. \textbf{Bold} indicates values at or above 95\% of the best performance.}
    \label{table:analysis_and_ablation_studies} 
    \vspace{-0.15in}
\end{table*}
\subsection{Ablation Studies and Analyses}
\label{sec:ablation}

    We investigate the effect of hyperparameters and various components of \metabbr by running experiments on OGBench puzzle-4x4 task.

    \vspace{-0.1in}
    \paragraph{Effect of action sequence length}
    Table~\ref{table:effect_of_as} investigates the impact of action sequence length on performance. When using single-step or two-step action ($H=1,2$), \metabbr fails to achieve meaningful performance, confirming the necessity of action sequences for long-horizon tasks. Performance improves with longer sequences, but when the sequence length becomes longer than 8, it requires proportionally larger actor networks to handle the increased action dimensions, suggesting a trade-off between sequence length and computational efficiency.
    \vspace{-0.1in}

    \paragraph{Effect of network size}
    Table~\ref{table:effect_of_critic_size} analyzes the sensitivity to network sizes. For the critic network, we observe that increasing capacity initially improves performance by better approximating the value function. For the value function, we find that the network needs sufficient capacity to capture the complexity of action sequence values, but excessive capacity without proper regularization causes instability in value estimation, leading to performance degradation.
    \vspace{-0.1in}

    \paragraph{Effect of training objective}
    In Table~\ref{table:objectives}, we compare different training objectives for value estimation. We found that using only distributional RL (HLG)~\citep{farebrother2024stop} or only standard regression (IQL) shows limited performance. However, combining detached value learning with distributional estimation significantly improves results, suggesting both components are crucial for stable training with action sequences.
    \vspace{-0.1in}

    \paragraph{Effect of dual discount factors}
    Lastly, we examine the effect of dual discount factors on learning dynamics in Table~\ref{table:discount_factor}. Proper tuning of $\gamma_1$ (the discount factor for action sequences) is essential for performance, as it controls the temporal horizon for value estimation within sequences. In this paper, we use $\gamma_1 = 0.9$ for all experiments.
    \vspace{-0.05in}

\section{Conclusion}
    We present \metabbr, a simple yet effective offline RL method that leverages action sequences for scalable learning in complex tasks. By modeling temporally extended actions through the options framework, \metabbr achieves principled horizon reduction via SMDP Q-learning while addressing value overestimation through detached value learning. Our experiments demonstrate consistent improvements over baselines on challenging OGBench tasks and successful application to large-scale VLAs, showing the practical potential for scaling offline RL to real-world scenarios.


\subsection*{Reproducibility statement}
We provide full hyperparameter and implementation details in \Cref{sec:exp} and \Cref{sec:implementation}. 
In addition, to further facilitate the reproduction, we release the open-sourced implementation through the \href{https://changyeon.site/deas}{project website}.
\subsubsection*{Acknowledgments}
CY thanks Jaehyun Nam, Juyong Lee, and anonymous reviewers for providing helpful feedback and suggestions for improving our paper. CY also thanks Angeline S. Kim for the assistance in improving the expression and the visualization of the paper.

\bibliography{iclr2026_conference}

\begin{thebibliography}{71}
\providecommand{\natexlab}[1]{#1}
\providecommand{\url}[1]{\texttt{#1}}
\expandafter\ifx\csname urlstyle\endcsname\relax
  \providecommand{\doi}[1]{doi: #1}\else
  \providecommand{\doi}{doi: \begingroup \urlstyle{rm}\Url}\fi

\bibitem[Ajay et~al.(2021)Ajay, Kumar, Agrawal, Levine, and Nachum]{ajay2021opal}
Anurag Ajay, Aviral Kumar, Pulkit Agrawal, Sergey Levine, and Ofir Nachum.
\newblock {\{}OPAL{\}}: Offline primitive discovery for accelerating offline reinforcement learning.
\newblock In \emph{International Conference on Learning Representations}, 2021.

\bibitem[Bacon et~al.(2017)Bacon, Harb, and Precup]{bacon2017option}
Pierre-Luc Bacon, Jean Harb, and Doina Precup.
\newblock The option-critic architecture.
\newblock In \emph{{AAAI} Conference on Artificial Intelligence}, 2017.

\bibitem[Baykal-G{\"u}rsoy \& G{\"u}rsoy(2010)Baykal-G{\"u}rsoy and G{\"u}rsoy]{baykal2010semi}
Melike Baykal-G{\"u}rsoy and K~G{\"u}rsoy.
\newblock Semi-markov decision processes.
\newblock \emph{Wiley Encyclopedia of Operations Research and Management Sciences}, 2010.

\bibitem[Bellemare et~al.(2017)Bellemare, Dabney, and Munos]{bellemare2017distributional}
Marc~G Bellemare, Will Dabney, and R{\'e}mi Munos.
\newblock A distributional perspective on reinforcement learning.
\newblock In \emph{International Conference on Machine Learning}, 2017.

\bibitem[Bjorck et~al.(2025)Bjorck, Casta{\~n}eda, Cherniadev, Da, Ding, Fan, Fang, Fox, Hu, Huang, et~al.]{bjorck2025gr00t}
Johan Bjorck, Fernando Casta{\~n}eda, Nikita Cherniadev, Xingye Da, Runyu Ding, Linxi Fan, Yu~Fang, Dieter Fox, Fengyuan Hu, Spencer Huang, et~al.
\newblock Gr00t n1: An open foundation model for generalist humanoid robots.
\newblock \emph{arXiv preprint arXiv:2503.14734}, 2025.

\bibitem[Black et~al.(2025)Black, Brown, Driess, Esmail, Equi, Finn, Fusai, Groom, Hausman, Ichter, et~al.]{black2025pi_0}
Kevin Black, Noah Brown, Danny Driess, Adnan Esmail, Michael Equi, Chelsea Finn, Niccolo Fusai, Lachy Groom, Karol Hausman, Brian Ichter, et~al.
\newblock $\pi_0$: A vision-language-action flow model for general robot control.
\newblock In \emph{Robotics: Science and Systems}, 2025.

\bibitem[Bradtke \& Duff(1994)Bradtke and Duff]{bradtke1994reinforcement}
Steven Bradtke and Michael Duff.
\newblock Reinforcement learning methods for continuous-time markov decision problems.
\newblock In \emph{Conference on Neural Information Processing Systems}, 1994.

\bibitem[Chebotar et~al.(2023)Chebotar, Vuong, Hausman, Xia, Lu, Irpan, Kumar, Yu, Herzog, Pertsch, et~al.]{chebotar2023q}
Yevgen Chebotar, Quan Vuong, Karol Hausman, Fei Xia, Yao Lu, Alex Irpan, Aviral Kumar, Tianhe Yu, Alexander Herzog, Karl Pertsch, et~al.
\newblock Q-transformer: Scalable offline reinforcement learning via autoregressive q-functions.
\newblock In \emph{Conference on Robot Learning}, 2023.

\bibitem[Chen et~al.(2023)Chen, Lu, Ying, Su, and Zhu]{chen2023offline}
Huayu Chen, Cheng Lu, Chengyang Ying, Hang Su, and Jun Zhu.
\newblock Offline reinforcement learning via high-fidelity generative behavior modeling.
\newblock In \emph{International Conference on Learning Representations}, 2023.

\bibitem[Chen et~al.(2025{\natexlab{a}})Chen, Tian, Liu, Zhou, Li, and Zhao]{chen2025conrft}
Yuhui Chen, Shuai Tian, Shugao Liu, Yingting Zhou, Haoran Li, and Dongbin Zhao.
\newblock Conrft: A reinforced fine-tuning method for vla models via consistency policy.
\newblock \emph{arXiv preprint arXiv:2502.05450}, 2025{\natexlab{a}}.

\bibitem[Chen et~al.(2025{\natexlab{b}})Chen, Niu, Kong, and Wang]{chen2025tgrpo}
Zengjue Chen, Runliang Niu, He~Kong, and Qi~Wang.
\newblock Tgrpo: Fine-tuning vision-language-action model via trajectory-wise group relative policy optimization.
\newblock \emph{arXiv preprint arXiv:2506.08440}, 2025{\natexlab{b}}.

\bibitem[Chi et~al.(2023)Chi, Xu, Feng, Cousineau, Du, Burchfiel, Tedrake, and Song]{chi2023diffusion}
Cheng Chi, Zhenjia Xu, Siyuan Feng, Eric Cousineau, Yilun Du, Benjamin Burchfiel, Russ Tedrake, and Shuran Song.
\newblock Diffusion policy: Visuomotor policy learning via action diffusion.
\newblock \emph{International Journal of Robotics Research}, 2023.

\bibitem[Farebrother et~al.(2024)Farebrother, Orbay, Vuong, Taiga, Chebotar, Xiao, Irpan, Levine, Castro, Faust, Kumar, and Agarwal]{farebrother2024stop}
Jesse Farebrother, Jordi Orbay, Quan Vuong, Adrien~Ali Taiga, Yevgen Chebotar, Ted Xiao, Alex Irpan, Sergey Levine, Pablo~Samuel Castro, Aleksandra Faust, Aviral Kumar, and Rishabh Agarwal.
\newblock Stop regressing: Training value functions via classification for scalable deep {RL}.
\newblock In \emph{International Conference on Machine Learning}, 2024.

\bibitem[Feinberg(1994)]{feinberg1994constrained}
Eugene~A Feinberg.
\newblock Constrained semi-markov decision processes with average rewards.
\newblock \emph{Zeitschrift f{\"u}r Operations Research}, 1994.

\bibitem[Fu et~al.(2020)Fu, Kumar, Nachum, Tucker, and Levine]{fu2020d4rl}
Justin Fu, Aviral Kumar, Ofir Nachum, George Tucker, and Sergey Levine.
\newblock D4rl: Datasets for deep data-driven reinforcement learning.
\newblock \emph{arXiv preprint arXiv:2004.07219}, 2020.

\bibitem[Fujimoto \& Gu(2021)Fujimoto and Gu]{fujimoto2021minimalist}
Scott Fujimoto and Shixiang~Shane Gu.
\newblock A minimalist approach to offline reinforcement learning.
\newblock In \emph{Conference on Neural Information Processing Systems}, 2021.

\bibitem[Fujimoto et~al.(2019)Fujimoto, Meger, and Precup]{fujimoto2019off}
Scott Fujimoto, David Meger, and Doina Precup.
\newblock Off-policy deep reinforcement learning without exploration.
\newblock In \emph{International Conference on Machine Learning}, 2019.

\bibitem[Garg et~al.(2023)Garg, Hejna, Geist, and Ermon]{garg2023extreme}
Divyansh Garg, Joey Hejna, Matthieu Geist, and Stefano Ermon.
\newblock Extreme q-learning: Maxent {RL} without entropy.
\newblock In \emph{International Conference on Learning Representations}, 2023.

\bibitem[Gulcehre et~al.(2020)Gulcehre, Wang, Novikov, Paine, G{\'o}mez, Zolna, Agarwal, Merel, Mankowitz, Paduraru, et~al.]{gulcehre2020rl}
Caglar Gulcehre, Ziyu Wang, Alexander Novikov, Thomas Paine, Sergio G{\'o}mez, Konrad Zolna, Rishabh Agarwal, Josh~S Merel, Daniel~J Mankowitz, Cosmin Paduraru, et~al.
\newblock Rl unplugged: A suite of benchmarks for offline reinforcement learning.
\newblock \emph{Conference on Neural Information Processing Systems}, 2020.

\bibitem[Guo et~al.(2025)Guo, Zhang, Chen, Ji, Wang, Hu, and Chen]{guo2025online}
Yanjiang Guo, Jianke Zhang, Xiaoyu Chen, Xiang Ji, Yen-Jen Wang, Yucheng Hu, and Jianyu Chen.
\newblock Improving vision-language-action model with online reinforcement learning.
\newblock \emph{arXiv preprint arXiv:2501.16664}, 2025.

\bibitem[Hansen-Estruch et~al.(2023)Hansen-Estruch, Kostrikov, Janner, Kuba, and Levine]{hansen2023idql}
Philippe Hansen-Estruch, Ilya Kostrikov, Michael Janner, Jakub~Grudzien Kuba, and Sergey Levine.
\newblock Idql: Implicit q-learning as an actor-critic method with diffusion policies.
\newblock \emph{arXiv preprint arXiv:2304.10573}, 2023.

\bibitem[Hendrycks \& Gimpel(2016)Hendrycks and Gimpel]{hendrycks2016gaussian}
Dan Hendrycks and Kevin Gimpel.
\newblock Gaussian error linear units (gelus).
\newblock \emph{arXiv preprint arXiv:1606.08415}, 2016.

\bibitem[Huang et~al.(2025)Huang, Fang, Zhang, Li, Zhao, and Xia]{huang2025corft}
Dongchi Huang, Zhirui Fang, Tianle Zhang, Yihang Li, Lin Zhao, and Chunhe Xia.
\newblock Co-rft: Efficient fine-tuning of vision-language-action models through chunked offline reinforcement learning.
\newblock \emph{arXiv preprint arXiv:2508.02219}, 2025.

\bibitem[Intelligence et~al.(2025)Intelligence, Black, Brown, Darpinian, Dhabalia, Driess, Esmail, Equi, Finn, Fusai, Galliker, Ghosh, Groom, Hausman, Ichter, Jakubczak, Jones, Ke, LeBlanc, Levine, Li-Bell, Mothukuri, Nair, Pertsch, Ren, Shi, Smith, Springenberg, Stachowicz, Tanner, Vuong, Walke, Walling, Wang, Yu, and Zhilinsky]{intelligence2025pi05visionlanguageactionmodelopenworld}
Physical Intelligence, Kevin Black, Noah Brown, James Darpinian, Karan Dhabalia, Danny Driess, Adnan Esmail, Michael Equi, Chelsea Finn, Niccolo Fusai, Manuel~Y. Galliker, Dibya Ghosh, Lachy Groom, Karol Hausman, Brian Ichter, Szymon Jakubczak, Tim Jones, Liyiming Ke, Devin LeBlanc, Sergey Levine, Adrian Li-Bell, Mohith Mothukuri, Suraj Nair, Karl Pertsch, Allen~Z. Ren, Lucy~Xiaoyang Shi, Laura Smith, Jost~Tobias Springenberg, Kyle Stachowicz, James Tanner, Quan Vuong, Homer Walke, Anna Walling, Haohuan Wang, Lili Yu, and Ury Zhilinsky.
\newblock $\pi_{0.5}$: a vision-language-action model with open-world generalization.
\newblock \emph{arXiv preprint arXiv:2504.16054}, 2025.

\bibitem[Kearns \& Singh(2000)Kearns and Singh]{kearns2000bias}
Michael~J Kearns and Satinder Singh.
\newblock Bias-variance error bounds for temporal difference updates.
\newblock In \emph{Conference on Learning Theory}, 2000.

\bibitem[Kingma(2015)]{kingma2014adam}
Diederik~P Kingma.
\newblock Adam: A method for stochastic optimization.
\newblock In \emph{International Conference on Learning Representations}, 2015.

\bibitem[Kostrikov et~al.(2022)Kostrikov, Nair, and Levine]{kostrikov2022offline}
Ilya Kostrikov, Ashvin Nair, and Sergey Levine.
\newblock Offline reinforcement learning with implicit q-learning.
\newblock In \emph{International Conference on Learning Representations}, 2022.

\bibitem[Kulkarni et~al.(2016)Kulkarni, Narasimhan, Saeedi, and Tenenbaum]{kulkarni2016hierarchical}
Tejas~D Kulkarni, Karthik Narasimhan, Ardavan Saeedi, and Josh Tenenbaum.
\newblock Hierarchical deep reinforcement learning: Integrating temporal abstraction and intrinsic motivation.
\newblock In \emph{Conference on Neural Information Processing Systems}, 2016.

\bibitem[Kumar et~al.(2019)Kumar, Fu, Soh, Tucker, and Levine]{kumar2019stabilizing}
Aviral Kumar, Justin Fu, Matthew Soh, George Tucker, and Sergey Levine.
\newblock Stabilizing off-policy q-learning via bootstrapping error reduction.
\newblock In \emph{Conference on Neural Information Processing Systems}, 2019.

\bibitem[Kumar et~al.(2020)Kumar, Zhou, Tucker, and Levine]{kumar2020conservative}
Aviral Kumar, Aurick Zhou, George Tucker, and Sergey Levine.
\newblock Conservative q-learning for offline reinforcement learning.
\newblock In \emph{Conference on Neural Information Processing Systems}, 2020.

\bibitem[Kumar et~al.(2023{\natexlab{a}})Kumar, Agarwal, Geng, Tucker, and Levine]{kumar2023offline}
Aviral Kumar, Rishabh Agarwal, Xinyang Geng, George Tucker, and Sergey Levine.
\newblock Offline q-learning on diverse multi-task data both scales and generalizes.
\newblock In \emph{International Conference on Learning Representations}, 2023{\natexlab{a}}.

\bibitem[Kumar et~al.(2023{\natexlab{b}})Kumar, Singh, Ebert, Nakamoto, Yang, Finn, and Levine]{kumar2022pre}
Aviral Kumar, Anikait Singh, Frederik Ebert, Mitsuhiko Nakamoto, Yanlai Yang, Chelsea Finn, and Sergey Levine.
\newblock Pre-training for robots: Offline rl enables learning new tasks from a handful of trials.
\newblock In \emph{Robotics: Science and Systems}, 2023{\natexlab{b}}.

\bibitem[Lange et~al.(2012)Lange, Gabel, and Riedmiller]{lange2012batch}
Sascha Lange, Thomas Gabel, and Martin Riedmiller.
\newblock Batch reinforcement learning.
\newblock In \emph{Reinforcement learning: State-of-the-art}, 2012.

\bibitem[Levine et~al.(2020)Levine, Kumar, Tucker, and Fu]{levine2020offline}
Sergey Levine, Aviral Kumar, George Tucker, and Justin Fu.
\newblock Offline reinforcement learning: Tutorial, review, and perspectives on open problems.
\newblock \emph{arXiv preprint arXiv:2005.01643}, 2020.

\bibitem[Li et~al.(2024)Li, Tian, Zhou, Jiang, Lioutikov, and Neumann]{li2024top}
Ge~Li, Dong Tian, Hongyi Zhou, Xinkai Jiang, Rudolf Lioutikov, and Gerhard Neumann.
\newblock Top-erl: Transformer-based off-policy episodic reinforcement learning.
\newblock \emph{arXiv preprint arXiv:2410.09536}, 2024.

\bibitem[Li et~al.(2025{\natexlab{a}})Li, Zuo, Yu, Zhang, Yang, Zhang, Zhu, Zhang, Chen, Cui, et~al.]{li2025simplevla}
Haozhan Li, Yuxin Zuo, Jiale Yu, Yuhao Zhang, Zhaohui Yang, Kaiyan Zhang, Xuekai Zhu, Yuchen Zhang, Tianxing Chen, Ganqu Cui, et~al.
\newblock Simplevla-rl: Scaling vla training via reinforcement learning.
\newblock \emph{arXiv preprint arXiv:2509.09674}, 2025{\natexlab{a}}.

\bibitem[Li et~al.(2025{\natexlab{b}})Li, Zhou, and Levine]{li2025reinforcement}
Qiyang Li, Zhiyuan Zhou, and Sergey Levine.
\newblock Reinforcement learning with action chunking.
\newblock \emph{arXiv preprint arXiv:2507.07969}, 2025{\natexlab{b}}.

\bibitem[Loshchilov \& Hutter(2019)Loshchilov and Hutter]{loshchilov2018decoupled}
Ilya Loshchilov and Frank Hutter.
\newblock Decoupled weight decay regularization.
\newblock In \emph{International Conference on Learning Representations}, 2019.

\bibitem[Mandlekar et~al.(2021)Mandlekar, Xu, Wong, Nasiriany, Wang, Kulkarni, Fei-Fei, Savarese, Zhu, and Mart{\'\i}n-Mart{\'\i}n]{mandlekar2021matters}
Ajay Mandlekar, Danfei Xu, Josiah Wong, Soroush Nasiriany, Chen Wang, Rohun Kulkarni, Li~Fei-Fei, Silvio Savarese, Yuke Zhu, and Roberto Mart{\'\i}n-Mart{\'\i}n.
\newblock What matters in learning from offline human demonstrations for robot manipulation.
\newblock \emph{arXiv preprint arXiv:2108.03298}, 2021.

\bibitem[Mandlekar et~al.(2023)Mandlekar, Nasiriany, Wen, Akinola, Narang, Fan, Zhu, and Fox]{mandlekar2023mimicgen}
Ajay Mandlekar, Soroush Nasiriany, Bowen Wen, Iretiayo Akinola, Yashraj Narang, Linxi Fan, Yuke Zhu, and Dieter Fox.
\newblock Mimicgen: A data generation system for scalable robot learning using human demonstrations.
\newblock In \emph{Conference on Robot Learning}, 2023.

\bibitem[Nachum et~al.(2018)Nachum, Gu, Lee, and Levine]{nachum2018data}
Ofir Nachum, Shixiang~Shane Gu, Honglak Lee, and Sergey Levine.
\newblock Data-efficient hierarchical reinforcement learning.
\newblock In \emph{Conference on Neural Information Processing Systems}, 2018.

\bibitem[Nair et~al.(2020)Nair, Gupta, Dalal, and Levine]{nair2020awac}
Ashvin Nair, Abhishek Gupta, Murtaza Dalal, and Sergey Levine.
\newblock Awac: Accelerating online reinforcement learning with offline datasets.
\newblock \emph{arXiv preprint arXiv:2006.09359}, 2020.

\bibitem[Nakamoto et~al.(2023)Nakamoto, Zhai, Singh, Sobol~Mark, Ma, Finn, Kumar, and Levine]{nakamoto2023cal}
Mitsuhiko Nakamoto, Simon Zhai, Anikait Singh, Max Sobol~Mark, Yi~Ma, Chelsea Finn, Aviral Kumar, and Sergey Levine.
\newblock Cal-ql: Calibrated offline rl pre-training for efficient online fine-tuning.
\newblock In \emph{Conference on Neural Information Processing Systems}, 2023.

\bibitem[Nakamoto et~al.(2024)Nakamoto, Mees, Kumar, and Levine]{nakamoto2024steering}
Mitsuhiko Nakamoto, Oier Mees, Aviral Kumar, and Sergey Levine.
\newblock Steering your generalists: Improving robotic foundation models via value guidance.
\newblock In \emph{Conference on Robot Learning}, 2024.

\bibitem[Nasiriany et~al.(2024)Nasiriany, Maddukuri, Zhang, Parikh, Lo, Joshi, Mandlekar, and Zhu]{nasiriany2024robocasa}
Soroush Nasiriany, Abhiram Maddukuri, Lance Zhang, Adeet Parikh, Aaron Lo, Abhishek Joshi, Ajay Mandlekar, and Yuke Zhu.
\newblock Robocasa: Large-scale simulation of everyday tasks for generalist robots.
\newblock In \emph{Robotics: Science and Systems}, 2024.

\bibitem[Nauman et~al.(2024)Nauman, Ostaszewski, Jankowski, Mi{\l}o{\'s}, and Cygan]{nauman2024bigger}
Michal Nauman, Mateusz Ostaszewski, Krzysztof Jankowski, Piotr Mi{\l}o{\'s}, and Marek Cygan.
\newblock Bigger, regularized, optimistic: scaling for compute and sample efficient continuous control.
\newblock In \emph{Conference on Neural Information Processing Systems}, 2024.

\bibitem[NVIDIA(2025)]{gr00tn1_5}
NVIDIA.
\newblock Gr00t n1.5: An improved open foundation model for generalist humanoid robots.
\newblock \url{https://research.nvidia.com/labs/gear/gr00t-n1_5/}, June 2025.
\newblock Accessed: 2025-09-09.

\bibitem[Oh et~al.(2018)Oh, Guo, Singh, and Lee]{oh2018self}
Junhyuk Oh, Yijie Guo, Satinder Singh, and Honglak Lee.
\newblock Self-imitation learning.
\newblock In \emph{International Conference on Machine Learning}, 2018.

\bibitem[Park et~al.(2024)Park, Frans, Levine, and Kumar]{park2024value}
Seohong Park, Kevin Frans, Sergey Levine, and Aviral Kumar.
\newblock Is value learning really the main bottleneck in offline rl?
\newblock In \emph{Conference on Neural Information Processing Systems}, 2024.

\bibitem[Park et~al.(2025{\natexlab{a}})Park, Frans, Eysenbach, and Levine]{park2025ogbench}
Seohong Park, Kevin Frans, Benjamin Eysenbach, and Sergey Levine.
\newblock {OGB}ench: Benchmarking offline goal-conditioned {RL}.
\newblock In \emph{International Conference on Learning Representations}, 2025{\natexlab{a}}.

\bibitem[Park et~al.(2025{\natexlab{b}})Park, Frans, Mann, Eysenbach, Kumar, and Levine]{park2025horizon}
Seohong Park, Kevin Frans, Deepinder Mann, Benjamin Eysenbach, Aviral Kumar, and Sergey Levine.
\newblock Horizon reduction makes rl scalable.
\newblock In \emph{Conference on Neural Information Processing Systems}, 2025{\natexlab{b}}.

\bibitem[Park et~al.(2025{\natexlab{c}})Park, Li, and Levine]{park2025flow}
Seohong Park, Qiyang Li, and Sergey Levine.
\newblock Flow q-learning.
\newblock In \emph{International Conference on Machine Learning}, 2025{\natexlab{c}}.

\bibitem[Peng et~al.(2019)Peng, Kumar, Zhang, and Levine]{peng2019advantage}
Xue~Bin Peng, Aviral Kumar, Grace Zhang, and Sergey Levine.
\newblock Advantage-weighted regression: Simple and scalable off-policy reinforcement learning.
\newblock \emph{arXiv preprint arXiv:1910.00177}, 2019.

\bibitem[Pomerleau(1988)]{pomerleau1988alvinn}
Dean~A Pomerleau.
\newblock Alvinn: An autonomous land vehicle in a neural network.
\newblock In \emph{Conference on Neural Information Processing Systems}, 1988.

\bibitem[Seo \& Abbeel(2025)Seo and Abbeel]{seo2024coarse}
Younggyo Seo and Pieter Abbeel.
\newblock Coarse-to-fine q-network with action sequence for data-efficient robot learning.
\newblock In \emph{Conference on Neural Information Processing Systems}, 2025.

\bibitem[Springenberg et~al.(2024)Springenberg, Abdolmaleki, Zhang, Groth, Bloesch, Lampe, Brakel, Bechtle, Kapturowski, Hafner, Heess, and Riedmiller]{springenberg2024offline}
Jost~Tobias Springenberg, Abbas Abdolmaleki, Jingwei Zhang, Oliver Groth, Michael Bloesch, Thomas Lampe, Philemon Brakel, Sarah Maria~Elisabeth Bechtle, Steven Kapturowski, Roland Hafner, Nicolas Heess, and Martin Riedmiller.
\newblock Offline actor-critic reinforcement learning scales to large models.
\newblock In \emph{International Conference on Machine Learning}, 2024.

\bibitem[Stolle \& Precup(2002)Stolle and Precup]{stolle2002learning}
Martin Stolle and Doina Precup.
\newblock Learning options in reinforcement learning.
\newblock In \emph{International Symposium on abstraction, reformulation, and approximation}, 2002.

\bibitem[Sutton \& Barto(2018)Sutton and Barto]{sutton2018reinforcement}
Richard~S Sutton and Andrew~G Barto.
\newblock \emph{Reinforcement learning: An introduction}.
\newblock MIT press, 2018.

\bibitem[Sutton et~al.(1999)Sutton, Precup, and Singh]{sutton1999between}
Richard~S Sutton, Doina Precup, and Satinder Singh.
\newblock Between mdps and semi-mdps: A framework for temporal abstraction in reinforcement learning.
\newblock \emph{Artificial intelligence}, 1999.

\bibitem[Tan et~al.(2025)Tan, Dou, Zhao, and Kr{\"a}henb{\"u}hl]{tan2025riptvla}
Shuhan Tan, Kairan Dou, Yue Zhao, and Philipp Kr{\"a}henb{\"u}hl.
\newblock Ript-vla: Interactive post-training for vision-language-action models.
\newblock \emph{arXiv preprint arXiv:2505.17016}, 2025.

\bibitem[Tarasov et~al.(2023)Tarasov, Kurenkov, Nikulin, and Kolesnikov]{tarasov2023revisiting}
Denis Tarasov, Vladislav Kurenkov, Alexander Nikulin, and Sergey Kolesnikov.
\newblock Revisiting the minimalist approach to offline reinforcement learning.
\newblock In \emph{Conference on Neural Information Processing Systems}, 2023.

\bibitem[Tian et~al.(2025)Tian, Li, Zhou, Celik, and Neumann]{tian2025chunking}
Dong Tian, Ge~Li, Hongyi Zhou, Onur Celik, and Gerhard Neumann.
\newblock Chunking the critic: A transformer-based soft actor-critic with n-step returns.
\newblock \emph{arXiv preprint arXiv:2503.03660}, 2025.

\bibitem[Tsitsiklis \& Van~Roy(1996)Tsitsiklis and Van~Roy]{tsitsiklis1996analysis}
John Tsitsiklis and Benjamin Van~Roy.
\newblock Analysis of temporal-diffference learning with function approximation.
\newblock In \emph{Conference on Neural Information Processing Systems}, 1996.

\bibitem[Vezhnevets et~al.(2017)Vezhnevets, Osindero, Schaul, Heess, Jaderberg, Silver, and Kavukcuoglu]{vezhnevets2017feudal}
Alexander~Sasha Vezhnevets, Simon Osindero, Tom Schaul, Nicolas Heess, Max Jaderberg, David Silver, and Koray Kavukcuoglu.
\newblock Feudal networks for hierarchical reinforcement learning.
\newblock In \emph{International Conference on Machine Learning}, 2017.

\bibitem[Wang et~al.(2023)Wang, Hunt, and Zhou]{wang2023diffusion}
Zhendong Wang, Jonathan~J Hunt, and Mingyuan Zhou.
\newblock Diffusion policies as an expressive policy class for offline reinforcement learning.
\newblock In \emph{International Conference on Learning Representations}, 2023.

\bibitem[Wang et~al.(2020)Wang, Novikov, Zolna, Merel, Springenberg, Reed, Shahriari, Siegel, Gulcehre, Heess, et~al.]{wang2020critic}
Ziyu Wang, Alexander Novikov, Konrad Zolna, Josh~S Merel, Jost~Tobias Springenberg, Scott~E Reed, Bobak Shahriari, Noah Siegel, Caglar Gulcehre, Nicolas Heess, et~al.
\newblock Critic regularized regression.
\newblock In \emph{Conference on Neural Information Processing Systems}, 2020.

\bibitem[Xu et~al.(2023)Xu, Jiang, Li, Yang, Wang, Chan, and Zhan]{xu2023offline}
Haoran Xu, Li~Jiang, Jianxiong Li, Zhuoran Yang, Zhaoran Wang, Victor Wai~Kin Chan, and Xianyuan Zhan.
\newblock Offline {RL} with no {OOD} actions: In-sample learning via implicit value regularization.
\newblock In \emph{International Conference on Learning Representations}, 2023.

\bibitem[Yu et~al.(2020)Yu, Quillen, He, Julian, Hausman, Finn, and Levine]{yu2020meta}
Tianhe Yu, Deirdre Quillen, Zhanpeng He, Ryan Julian, Karol Hausman, Chelsea Finn, and Sergey Levine.
\newblock Meta-world: A benchmark and evaluation for multi-task and meta reinforcement learning.
\newblock In \emph{Conference on Robot Learning}, 2020.

\bibitem[Zhang et~al.(2025)Zhang, Zhuang, Zhao, Ding, Lu, and Wang]{zhang2025reinbot}
Hongyin Zhang, Zifeng Zhuang, Han Zhao, Pengxiang Ding, Hongchao Lu, and Donglin Wang.
\newblock Reinbot: Amplifying robot visual-language manipulation with reinforcement learning.
\newblock \emph{arXiv preprint arXiv:2505.07395}, 2025.

\bibitem[Zhang et~al.(2024)Zhang, Zheng, Chen, Jang, Li, Han, Wang, Ding, Fox, and Yao]{zhang2024grape}
Zijian Zhang, Kaiyuan Zheng, Zhaorun Chen, Joel Jang, Yi~Li, Siwei Han, Chaoqi Wang, Mingyu Ding, Dieter Fox, and Huaxiu Yao.
\newblock Grape: Generalizing robot policy via preference alignment.
\newblock \emph{arXiv preprint arXiv:2411.19309}, 2024.

\bibitem[Zhao et~al.(2023)Zhao, Kumar, Levine, and Finn]{zhao2023learning}
Tony~Z Zhao, Vikash Kumar, Sergey Levine, and Chelsea Finn.
\newblock Learning fine-grained bimanual manipulation with low-cost hardware.
\newblock In \emph{Robotics: Science and Systems}, 2023.

\end{thebibliography}
\bibliographystyle{iclr2026_conference}

\clearpage
\appendix

\section{Implementation and Training Details}
\label{sec:implementation}
    \subsection{OGBench Experiments}
    \label{sec:ogbench_implementation}
        \paragraph{Tasks} We evaluate our method on 6 UR5 Robot Arm manipulation environments from OGBench~\citep{park2025ogbench}, each with 5 subtasks. All tasks are state-based, and goal-free setup. For each task, the observation space consists of the proprioceptive state of the UR5 Robot Arm, and low-dim state vector informing the target object state and position. The action space consists of the cartesian position of UR5 robot arm, gripper yaw, and gripper open/close. For substituting goal-conditioned environment to standard function, we use the simple semi-sparse reward function, which is defined as the negative number of uncompleted subtasks in the current state, following \citet{park2025ogbench}. For all tasks, the maximum episode length is set to 1000.

        \paragraph{Implementation details} We implement our method on top of the open-source implementation of FQL~\citep{park2025flow}~\footnote{\url{https://github.com/seohongpark/fql}}. Unless otherwise mentioned, we largely follow the training/evaluation setup and network architecture from \citet{park2025flow} and \citet{park2025horizon}. For training value network, we use the smaller size network compared to critic network for all experiments, which shows the best performance, and we use the doubled size of network for the critic network. For $\tt{cube}$ experiments, we use BRO~\citep{nauman2024bigger} for additional regularization between relatively small range of returns in value function training. For selecting $\mathbf{v}_{\min}$ and $\mathbf{v}_{\max}$ for distributional RL, we use two procedures: 1) \textit{data-centric}: compute return distribution from the dataset and select 1\% and 99\% quantiles with 20\% padding, and 2) \textit{universal}: compute theoretical bounds using reward range $[r_{\min}, r_{\max}]$, horizon $L$, option length $H$, and discount factors $\gamma_1, \gamma_2$. For SMDP with $K = L/H$ options, the theoretical bounds are:
        \begin{align}
        \mathbf{v}_{\min} &= r_{\min} \frac{1 - \gamma_2^H}{1 - \gamma_2} \frac{1 - \gamma_1^K}{1 - \gamma_1} \\
        \mathbf{v}_{\max} &= r_{\max} \frac{1 - \gamma_2^H}{1 - \gamma_2} \frac{1 - \gamma_1^K}{1 - \gamma_1}
        \end{align}
        where $\gamma_1$ and $\gamma_2$ denote option-level and action-level discount factors, respectively.

        \paragraph{Training and evaluation} For the training dataset, we use the open-sourced 1M/100M $\tt{play}$ dataset released by \citet{park2025ogbench}~\footnote{\url{https://github.com/seohongpark/ogbench}}, where the dataset is collected by open-loop, non-Markovian scripted policies with temporally correlated noise. As 100M dataset consists of 100 separate files with 1M transitions for each, we use the first 10 files sorted by name for 10M dataset. We train our method and baselines for 1M (1M data) / 2.5M (10M/100M data) gradient steps. For selecting BC coefficient $\alpha$ for policy extraction, we first normalize the Q loss as in \citet{fujimoto2021minimalist} and sweep the value from $\{0.1, 0.3, 1, 3, 10\}$ and choose the best one for each task and baseline, except $\mathtt{cube-double}$, where we follow the hyperparameter used in \citet{li2025reinforcement}. For evaluation, we report the average success rates across the last three evaluation epochs (800K, 900K, 1M for 1M dataset, 2.3M, 2.4M, 2.5M for 10M/100M dataset) following \citet{park2025flow} and \citet{park2025horizon}. For checking additional hyperparameters used in our experiments, please refer to~\Cref{sec:hyp}.

        \paragraph{Baselines} For reporting results from FQL and $n$-step FQL, we use the implementation from \citet{park2025flow}. For Q-Chunking, we re-implement the code from \citet{li2025reinforcement}~\footnote{\url{https://github.com/ColinQiyangLi/qc}} in our codebase. We found that simply increasing discount factor $\gamma$ leads to significant performance improvement for Q-Chunking, so we use the discount factor to be same with $\gamma_2$ for value function training. For implementing CQN-AS, we use the original implementation released by the authors from \citet{seo2024coarse}~\footnote{\url{https://github.com/younggyoseo/CQN-AS}} and integrate OGBench related codes on top of the codebase. Originally, CQN-AS is designed to apply auxiliary BC loss only on expert demonstrations, but considering the dataset distribution of OGBench tasks with nearly no success rollouts, we modify the BC loss on the suboptimal data as well~\citep{fujimoto2021minimalist,park2025flow,park2024value}, where no significant difference with the original implementation. As the reward scale for OGBench is highly different according to the domain, we normalize the reward scale to be in $[-1,0]$, and use $\mathbf{v}_{\min}$ and $\mathbf{v}_{\max}$ as $-200$ and $0$, respectively. For levels and bins, we use 5 (level) and 9 (bins) for all experiments.

        \paragraph{Computing hardware} For all OGBench experiments, we use a single NVIDIA RTX 3090 GPU with 24GB VRAM and it takes about 2 hours for training the small model (used for 1M dataset) and about 8 hours for training the large model (used for 10M/100M dataset). 

    \subsection{VLA Experiments}
        \label{sec:vla_exp}
        \paragraph{Computing hardware} For all VLA experiments, we use NVIDIA A100 80GB GPUs. Fine-tuning GR00T N1.5 takes about 4 hours for 100 expert demonstrations and successful rollouts. For training \metabbr and baselines, it takes about 10 hours with the same data, as we use a larger batch size.

        \paragraph{VLA fine-tuning} We implement our method and baselines on top of the open-source implementation of GR00T N1.5~\citep{gr00tn1_5}~\footnote{\url{https://github.com/NVIDIA/Isaac-GR00T}}. As our code is based on an earlier version of GR00T N1.5, we conduct experiments without introducing future tokens to the action expert modules. For fine-tuning GR00T N1.5, we use a batch size of 32 and train for 30K (RoboCasa Kitchen) / 10K (Real Robot) steps using AdamW~\citep{loshchilov2018decoupled} optimizer with learning rate $1 \times 10^{-4}$ and cosine annealing schedule. 

        \subsubsection{RoboCasa Kitchen Experiments}
        \paragraph{Task} RoboCasa Kitchen~\citep{nasiriany2024robocasa} is a simulation environment with a mobile manipulator attached to a Franka Panda robot arm in household kitchen environments. Among 24 atomic tasks provided by the environment, we select 4 challenging tasks ($\tt{CoffeeSetupMug}$, $\tt{PnPMicrowaveToCounter}$, $\tt{PnPMicrowaveToMicrowave}$, $\tt{PnPMicrowaveToStove}$) that require relatively long-horizon and delicate manipulation with small grasping part, which is demonstrated by the low success rate of the base model. For perception, camera images from 3 different viewpoints (left front, right front, wrist), proprioceptive states including position/velocities of joint/base, and natural language instructions, are provided. For reward function, we use the pre-defined success detector in the environment, and use the sparse reward function where the reward is $1$ if the task is completed, and $0$ otherwise.

        \paragraph{Implementation details} As an input for the value function, we first use the proprioceptive states from the robot, including joint position/angle, base position/orientation for the mobile manipulator. To provide information on target objects to the value function, we utilize the encoded representation of three different camera views and task instructions from the VLM backbone. For the value/critic network architecture, we use the same hyperparameters as those used for the 100M dataset experiments. For optimizing value and critic function, we use the expectile parameter $\tau$ as 0.7, and use $\gamma_1 = 0.9$, $\gamma_2 = 0.99$, \textit{universal} support type for distributional RL, for all experiments. For selecting action candidates with the value function, we first sample $N=10$ candidates from the policy. For selecting final actions, we try either 1) greedy sampling with highest Q-value or 2) inspired by \citet{nakamoto2024steering}, sampling the action from a categorical distribution obtained by temperature controlled softmax over Q-values: $a_t \sim \text{Softmax}(\frac{Q(s_t, a_1)}{\beta},\ldots, \frac{Q(s_t, a_N)}{\beta})$ with temperature $\beta=1$ and report the best result for each task.  

        \paragraph{Training and evaluation} For expert demonstrations, we randomly sample 100 expert demonstrations using the publicly available dataset generated by MimicGen~\citep{mandlekar2023mimicgen}. For training \metabbr and baselines, we use a batch size of 64 and train for 30K steps using Adam optimizer with a learning rate of $3 \times 10^{-4}$. For collecting rollouts, we use randomized environments using the object instance set $A$. For each task, we evaluate the model performance across 50 trials on five distinct evaluation scenes with 3 different evaluation seeds, totaling 150 rollouts. To test generalization capabilities, we evaluate the policy only on unseen object instances.

        \subsubsection{Real Robot Experiments}
        \label{sec:real_robot_exp}
\begin{figure}
    \centering
    \includegraphics[width=\textwidth]{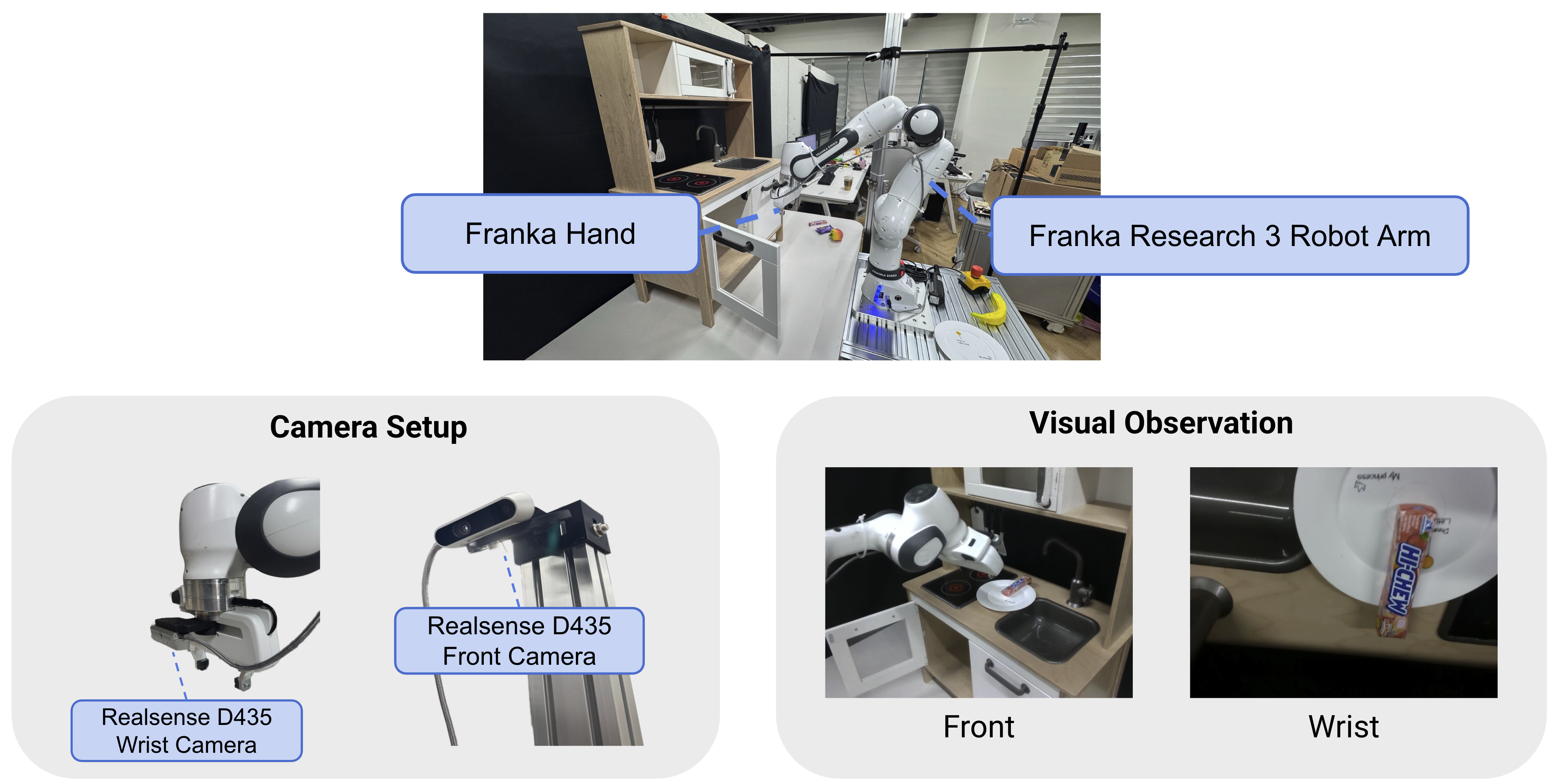}
    \caption{
        \textbf{Real-robot platform.} } 
    \label{fig:real_setup}
\end{figure}

        \paragraph{Hardware platform} We use Franka Research 3, a 7-DoF robotic arm, for our experiments. For visual perception, we utilize the dual camera with Intel RealSense D435i: a camera attached to the column next to the robot base to provide a global view, and a wrist-mounted camera for a close-range view. Teleoperated demonstrations are collected using an Oculus Quest 2, and we log time-synchronized RGB images, joint states, and gripper width for data collection. Demonstrations are recorded at 15 Hz. See \Cref{fig:real_setup} for visual examples.

\begin{figure}
    \centering
    \begin{subfigure}[b]{0.19\textwidth}
        \centering
        \includegraphics[width=\textwidth]{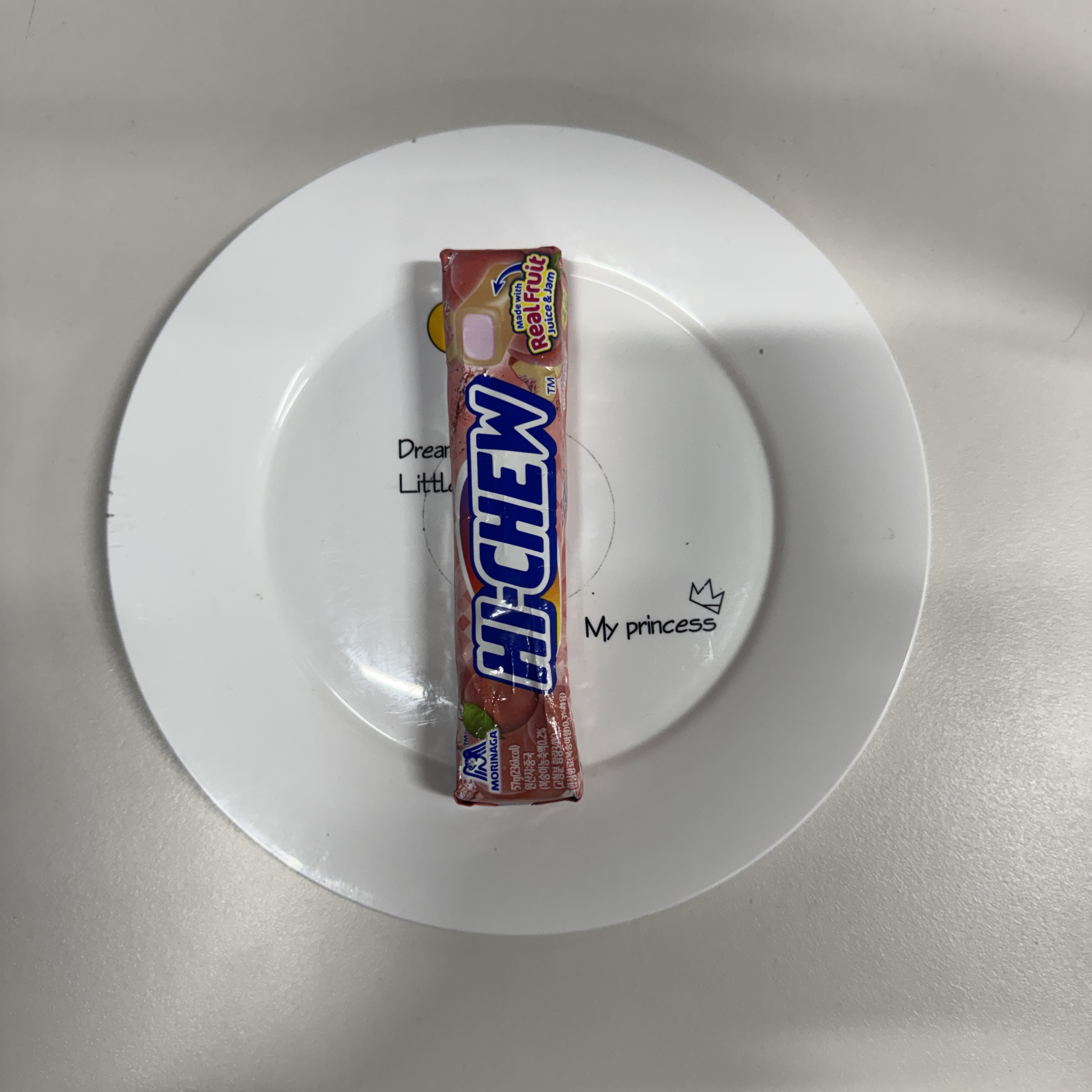}
        \label{fig:fr3_init1}
    \end{subfigure}
    \begin{subfigure}[b]{0.19\textwidth}
        \centering
        \includegraphics[width=\textwidth]{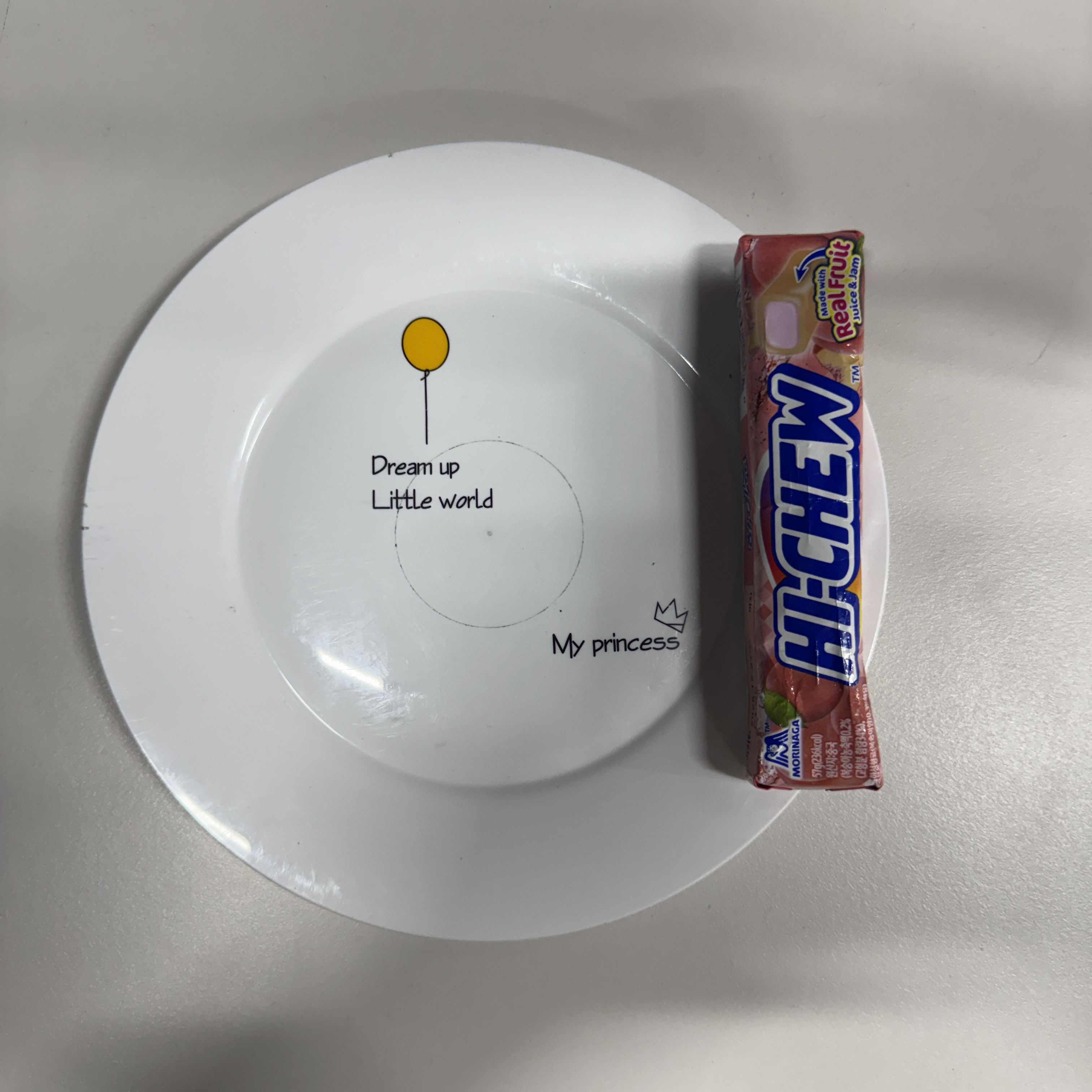}
        \label{fig:fr3_init2}
    \end{subfigure}
     \begin{subfigure}[b]{0.19\textwidth}
        \centering
        \includegraphics[width=\textwidth]{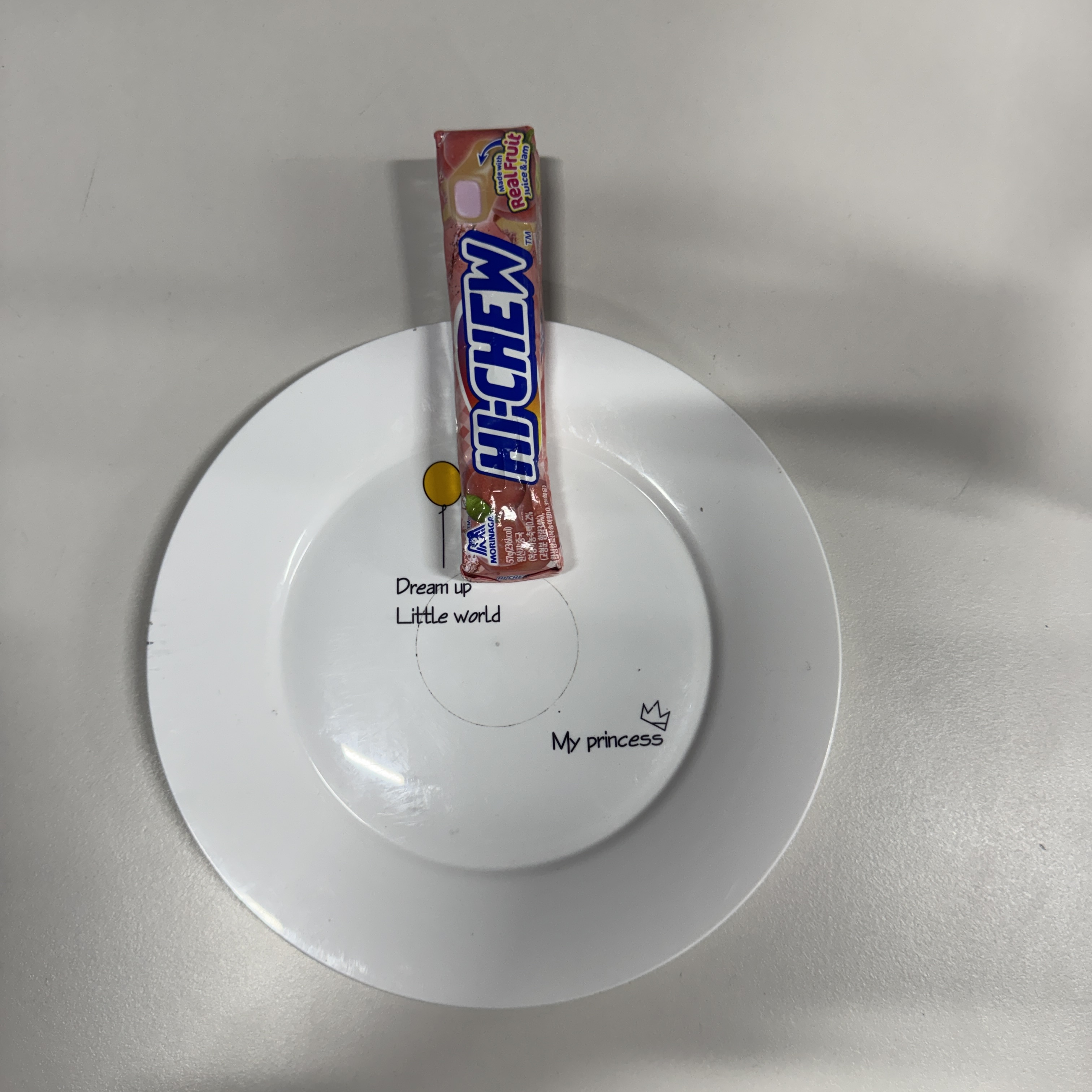}
        \label{fig:fr3_init3}
    \end{subfigure}
     \begin{subfigure}[b]{0.19\textwidth}
        \centering
        \includegraphics[width=\textwidth]{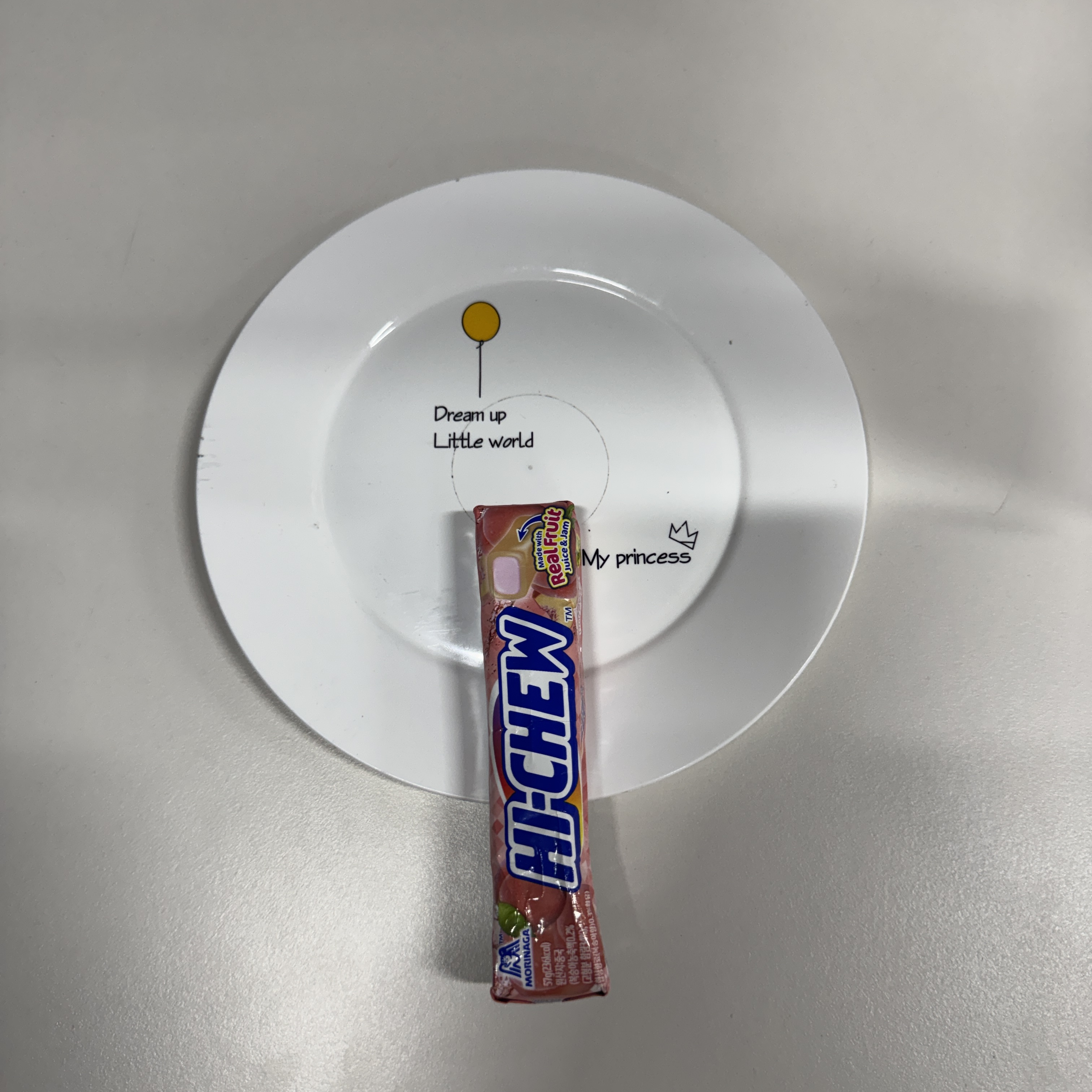}
        \label{fig:fr3_init4}
    \end{subfigure}
    \begin{subfigure}[b]{0.19\textwidth}
        \centering
        \includegraphics[width=\textwidth]{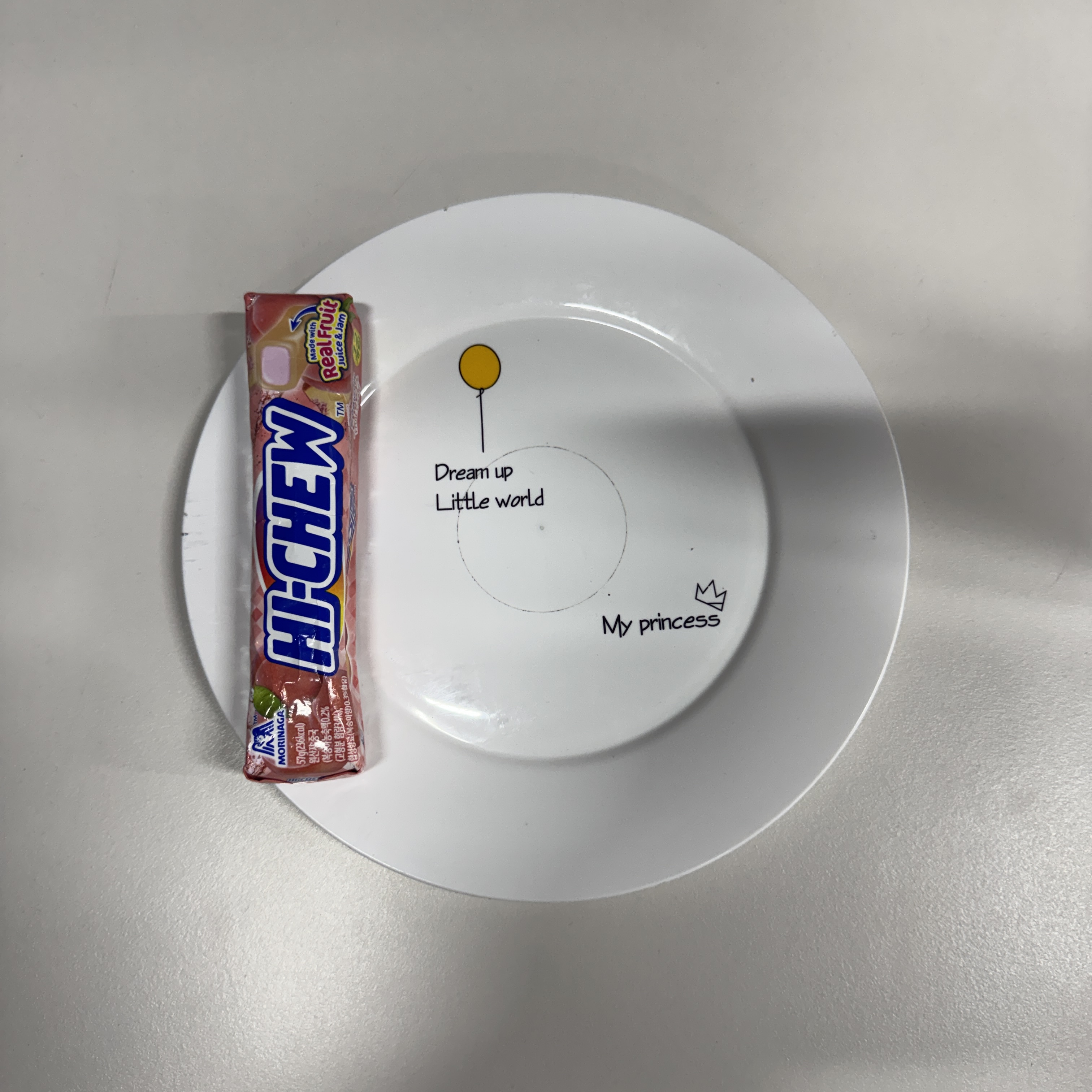}
        \label{fig:fr3_init5}
    \end{subfigure}
    \vspace{-0.15in}
    \caption{
        \textbf{Initialization points used for pick-and-place tasks.} } 
    \label{fig:fr3_init}
    \vspace{-0.1in}
\end{figure}

        \paragraph{Task} We evaluate the model performance on pick-and-place tasks from the countertop to the bottom cabinet, with three different objects: $\tt{peach}$, $\tt{milka}$, and $\tt{hichew}$. Each object has different properties: $\tt{peach}$ is a rigid object with a relatively larger size that is easy to occlude, $\tt{milka}$ is a deformable object with a relatively smaller size that is easy to deform, and $\tt{hichew}$ is a hard object requiring precise grasping due to its small width. For collecting demonstrations, we use different initialization points (center, top, bottom, left, right) and collect one demonstration for each position (see \Cref{fig:fr3_init} for the initialization points used in our experiments). For accurate value function estimation, we manually label the reward function for each task. Specifically, we split the task into 4 stages: 1) moving to the countertop, 2) picking up the object, 3) moving to the target position, and 4) placing the object. For each stage, we label the reward function as $1$ if the task is completed, and $0$ otherwise, and we set the reward function as the negative number of uncompleted stages following \citet{park2025ogbench}.

        \paragraph{Implementation details} Unless otherwise mentioned, we follow the same implementation details as in the RoboCasa Kitchen experiments. For selecting final actions, we use $N=50$ candidates from the policy and use the same procedure for selecting the final action as in the RoboCasa Kitchen experiments.
    \clearpage
    \subsection{Hyperparameters} 
    \label{sec:hyp}
    We list the hyperparameters used in our OGBench experiments in \Cref{tab:hyp,tab:hyp_baseline}. For the BC coefficient $\alpha$ used for policy extraction, please refer to \Cref{tab:policy_hyp}. 

        \begin{table}[h!]
            \caption{
            \footnotesize
            \textbf{\metabbr hyperparameters for OGBench experiments.}
            }
            \label{tab:hyp}
            \begin{center}
            \resizebox{0.8\textwidth}{!}{
            {
            \begin{tabular}{ll}
                \toprule
                \textbf{Hyperparameter} & \textbf{Value} \\
                \midrule
                Gradient steps & $1$M (1M dataset), $2.5$M (10M/100M dataset) \\
                Optimizer & Adam~\citep{kingma2014adam} \\
                Learning rate & $0.0003$ \\
                Batch size & $256$ (1M dataset), $1024$ (10M/100M dataset) \\
                Actor MLP size & $[512, 512, 512, 512]$ (1M dataset)\\
                & $[1024, 1024, 1024, 1024]$ (10M/100M dataset) \\
                Critic MLP size & $[256, 256, 256, 256]$ (1M dataset)\\
                & $[512, 512, 512, 512]$ (10M/100M dataset) \\
                Value MLP size & $[128, 128, 128, 128]$ (1M dataset)\\
                & $[256, 256, 256, 256]$ (10M/100M dataset) \\
                Nonlinearity & GELU~\citep{hendrycks2016gaussian} \\
                Layer normalization & True \\
                Target network update rate & $0.005$ \\
                Discount factor $\gamma_1$ & $0.9$ \\
                Discount factor $\gamma_2$ & $0.995$ ($\tt{cube}$), $0.999$ ($\tt{scene}, \tt{puzzle}$) \\
                HL-Gaussian - Atoms & 101 \\
                HL-Gaussian - $\sigma$ & 0.75 \\
                HL-Gaussian - Support range type & \textit{data-centric} ($\tt{cube}$), \textit{universal} ($\tt{scene}, \tt{puzzle}$) \\
                Flow steps & $10$ \\
                Critic ensemble size & $2$ \\
                Action sequence length $H$ & 4 ($\tt{cube}$), 8 (\tt{scene}, \tt{puzzle}) \\
                Expectile $\kappa$ (\metabbr) & $0.9$ (1M dataset), $0.95$ (10M/100M dataset) \\
                Double Q aggregation & $\min(Q_1, Q_2)$ \\
                Policy extraction hyperparameters & \Cref{tab:policy_hyp} \\
                \bottomrule
            \end{tabular}
            }
            }
            \end{center}
        \end{table} 

        \begin{table}[h!]
                \caption{
                \footnotesize
                \textbf{Baseline hyperparameters for OGBench experiments.} 
                }
                \label{tab:hyp_baseline}
                \begin{center}
                \resizebox{0.8\textwidth}{!}{
                {
                \begin{tabular}{ll}
                    \toprule
                    \textbf{Hyperparameter} & \textbf{Value} \\
                    \midrule
                    Critic MLP size & $[512, 512, 512, 512]$ (1M dataset)\\
                    & $[1024, 1024, 1024, 1024]$ (10M/100M dataset) \\
                    Discount factor $\gamma$ (FQL, $n$-step FQL) & $0.99$ \\
                    Discount factor $\gamma$ (QC-FQL) & $0.995$ (\tt{cube}), $0.999$ (\tt{puzzle}) \\
                    Horizon reduction factor $n$ & $4$ (\tt{cube}), $8$ (\tt{puzzle}) \\
                    Policy extraction hyperparameters & \Cref{tab:policy_hyp} \\
                    \midrule
                    Levels (CQN-AS) & 5 \\
                    Bins (CQN-AS) & 9 \\
                    C51 - $\mathbf{v}_{\text{min}}, \mathbf{v}_{\text{max}}$ (CQN-AS) & -200, 0 \\
                    \bottomrule
                \end{tabular}
                }
                }
                \end{center}
        \end{table} 
        
        \begin{table}[h!]
        \caption{
        \footnotesize
        \textbf{Policy extraction hyperparameters for OGBench experiments.} Note that we apply Q-Normalization~\citep{fujimoto2021minimalist} for actor loss, except \tt{cube-double} tasks.
        }
        \label{tab:policy_hyp}
        \begin{center}
        \resizebox{0.8\textwidth}{!}{
        {
        \begin{tabular}{lcccc}
            \toprule
            \tt{Task} & FQL $\alpha$ & $n$-step FQL $\alpha$ & QC-FQL $\alpha$ & \metabbr $\alpha$ \\
            \midrule
            \tt{scene} & $3$ & $1$ & $3$ & $3$ \\
            \tt{cube-double} & $300$ & $100$ & $300$ & $300.0$ \\
            \tt{puzzle-3x3} & $3$ & $1$ & $1$ & $3$ \\
            \tt{cube-triple} & $3$ & $1$ & $1$ & $1$ \\
            \tt{puzzle-4x4} & $3$ & $1$ & $1$ & $3$ \\
            \tt{cube-quadruple} & $3$ & $1$ & $1$ & $1$ \\
            \bottomrule
        \end{tabular}
        }
        }
        \end{center}
        \end{table}

\section{Extended Related Work}
\paragraph{Hierarchical RL and Options Framework}
Some Hierarchical RL works seek to address the challenges of long-horizon and sparse-reward tasks by reducing the 
effective horizon through learning value functions that consume multi-step actions~\citep{kulkarni2016hierarchical,
vezhnevets2017feudal,nachum2018data,ajay2021opal}, usually combined with bi-level architectures. Among them, Options 
framework~\citep{sutton1999between, stolle2002learning,bacon2017option} introduces formalization of higher-level actions 
that persist for multiple time steps with variable initiation/termination conditions, effectively reducing the planning 
horizon and facilitating more efficient learning. Our approach leverages the options perspective by treating action 
sequences as primitive options, enabling horizon reduction and improved value propagation without task-specific knowledge, 
explicit goal conditioning, or manual sub-task specification.

\paragraph{Reinforcement learning with VLAs} Recent efforts have applied RL to VLA training~\citep{zhang2024grape,chen2025conrft,zhang2025reinbot,guo2025online,tan2025riptvla,chen2025tgrpo,li2025simplevla}, but most focus on on-policy online RL, which requires expensive interactions and cannot reuse transitions. A key limitation is that existing methods use single-step value functions $Q(s,a)$ for value learning, despite modern VLAs being designed to predict action sequences~\citep{black2025pi_0,bjorck2025gr00t,intelligence2025pi05visionlanguageactionmodelopenworld}. This mismatch between single-step value learning and multi-step action prediction limits the effectiveness of RL with VLAs. The most related work is CO-RFT~\citep{huang2025corft}, which applies chunked offline RL to VLA training, but differs from our approach in three key aspects: (1) CO-RFT uses actor-critic methods~\citep{nakamoto2023cal} with single-step value functions while \metabbr uses detached value learning with action sequences, (2) CO-RFT relies on human teleoperated expert demonstrations while we use small expert sets with large suboptimal rollouts, and (3) CO-RFT requires sophisticated transformer architectures while \metabbr achieves improvements with simple MLP networks.

\section{Limitations and Future Work}
\label{sec:limitations}
While \metabbr demonstrates significant improvements over existing offline RL methods, several limitations and opportunities for future research remain. First, our current approach uses fixed action sequence lengths across different tasks, but the optimal sequence length varies significantly depending on task complexity. Future work should investigate adaptive mechanisms that can dynamically adjust action sequence lengths based on task requirements, potentially through adopting hierarchical policies~\citep{kulkarni2016hierarchical,vezhnevets2017feudal,nachum2018data}. Second, while \metabbr shows promising results on individual tasks, scaling to large-scale unified value functions remains a critical challenge for real-world deployment. \metabbr currently trains reward models on 3-4 tasks simultaneously, but practical applications require learning from hundreds or thousands of diverse tasks. Future research should focus on developing scalable architectures and training procedures that can handle massive multi-task datasets while maintaining sample efficiency and avoiding catastrophic forgetting. Third, our method relies on distributional RL with fixed support ranges ($\mathbf{v}_{\min}$, $\mathbf{v}_{\max}$) and discretization parameters, which can significantly impact performance. The sensitivity to these hyperparameters limits the method's robustness across different domains and reward scales. Future work should develop more robust frameworks that can automatically adapt to different reward distributions or provide principled ways to set these parameters.

\section{Use of Large Language Models}
We acknowledge the use of large language models (LLMs) in preparing this manuscript. LLMs were
employed solely to refine writing quality, including grammar correction, vocabulary suggestions, and
typographical checks. All substantive ideas, analyses, and conclusions in this paper are entirely the
work of the authors

\clearpage
\section{Full Experimental Results}
We include the full experimental results in OGBench experiments in Table~\ref{tab:full}.
\begin{table}[h]
    \caption{Full offline RL Results in \textbf{30} OGBench tasks. $^{*}$ indicates the default task in each environment. We report the success rate (\%) and 95\% stratified bootstrap confidence interval over 4 runs.}
    \centering
    \resizebox{\textwidth}{!}{
    \begin{tabular}{lccccc>{\columncolor{green!10}}c}
        \toprule
         \tt{Task} & \#Data & FQL & N-step FQL & QC-FQL & CQN-AS & \textbf{\metabbr} \\
         \midrule
         \tt{scene-play-singletask-task1-v0} & &  $\mathbf{100}$ {\tiny $\pm 0$} &  $\mathbf{100}$ {\tiny $\pm 0$}  &  $\mathbf{99}$ {\tiny $\pm 0$} & $2$ {\tiny $\pm 1$} & $\mathbf{99}$ {\tiny $\pm 1$}\\
         \tt{scene-play-singletask-task2-v0} & & $50$ {\tiny $\pm 7$}   & $4$ {\tiny $\pm 3$}  & $\mathbf{99}$ {\tiny $\pm 1$} & $1$ {\tiny $\pm 1$} & $\mathbf{97}$ {\tiny $\pm 1$}\\
         \tt{scene-play-singletask-task3-v0} & 1M &  $\mathbf{95}$ {\tiny $\pm 2$}   &  $78$ {\tiny $\pm 5$}  &  $64$ {\tiny $\pm 8$} & $0$ {\tiny $\pm 0$} & $75$ {\tiny $\pm 6$}\\
         \tt{scene-play-singletask-task4-v0$^{*}$} & & $3$ {\tiny $\pm 2$}   & $0$ {\tiny $\pm 0$}  &  $\mathbf{68}$ {\tiny $\pm 1$} & $0$ {\tiny $\pm 0$} & $\mathbf{65}$ {\tiny $\pm 5$}\\
         \tt{scene-play-singletask-task5-v0} & & $0$ {\tiny $\pm 0$}   & $0$ {\tiny $\pm 0$}  &  $35$ {\tiny $\pm 7$} & $0$ {\tiny $\pm 0$} & $\mathbf{45}$ {\tiny $\pm 6$}\\
         \midrule
         \tt{cube-double-play-singletask-task1-v0} & &  46 {\tiny $\pm 4$}   &  17 {\tiny $\pm 3$}  &  68 {\tiny $\pm 4$} & $7$ {\tiny $\pm 1$} & $\mathbf{76}$ {\tiny $\pm 3$}\\
         \tt{cube-double-play-singletask-task2-v0$^{*}$} & & $10$ {\tiny $\pm 2$}   & $1$ {\tiny $\pm 0$}  & $47$ {\tiny $\pm 8$} & $1$ {\tiny $\pm 1$} & $\mathbf{51}$ {\tiny $\pm 8$}\\
         \tt{cube-double-play-singletask-task3-v0} & 1M &  9 {\tiny $\pm 2$}   &  1 {\tiny $\pm 1$}  &  40{\tiny $\pm 6$} & $0$ {\tiny $\pm 1$} & $\mathbf{47}$ {\tiny $\pm 4$}\\
         \tt{cube-double-play-singletask-task4-v0} & &  1 {\tiny $\pm 1$}   &  0 {\tiny $\pm 0$}  &  $\mathbf{8}${\tiny $\pm 1$} & $1$ {\tiny $\pm 1$} & $\mathbf{8}$ {\tiny $\pm 1$}\\
         \tt{cube-double-play-singletask-task5-v0} & &  2 {\tiny $\pm 1$}   &  3 {\tiny $\pm 1$}  &  44{\tiny $\pm 3$} & $0$ {\tiny $\pm 0$} & $\mathbf{57}$ {\tiny $\pm 3$}\\
         \midrule
         \tt{puzzle-3x3-play-singletask-task1-v0} & &  $\mathbf{100}$ {\tiny $\pm 0$}   & $89$ {\tiny $\pm 3$}  &  $\mathbf{97}$ {\tiny $\pm 1$} & $1$ {\tiny $\pm 2$} & $\mathbf{100}$ {\tiny $\pm 0$}\\
         \tt{puzzle-3x3-play-singletask-task2-v0} & &  $19$ {\tiny $\pm 4$}   &  $40$ {\tiny $\pm 10$}  &  $81$ {\tiny $\pm 12$} & $0$ {\tiny $\pm 0$}& $\mathbf{94}$ {\tiny $\pm 5$}\\
         \tt{puzzle-3x3-play-singletask-task3-v0} & 1M &  $15$ {\tiny $\pm 2$}   &  $14$ {\tiny $\pm 3$}  &  $50$ {\tiny $\pm 11$} & $0$ {\tiny $\pm 0$}& $\mathbf{91}$ {\tiny $\pm 3$}\\
         \tt{puzzle-3x3-play-singletask-task4-v0$^{*}$} & & $35$ {\tiny $\pm 4$} & $23$ {\tiny $\pm 3$} & $31$ {\tiny $\pm 4$} & $0$ {\tiny $\pm 0$}& $\mathbf{91}$ {\tiny $\pm 3$}\\
         \tt{puzzle-3x3-play-singletask-task5-v0} & &  $47$ {\tiny $\pm 4$}   &  $13$ {\tiny $\pm 3$}  &  50 {\tiny $\pm 11$} & $0$ {\tiny $\pm 0$}& $\mathbf{96}$ {\tiny $\pm 2$}\\
         \midrule
         \tt{cube-triple-play-singletask-task1-v0} & &  31 {\tiny $\pm 14$}   &  17 {\tiny $\pm 5$}  &  $\mathbf{100}$ {\tiny $\pm 0$} & $0$ {\tiny $\pm 0$} & $\mathbf{98}$ {\tiny $\pm 1$}\\
         \tt{cube-triple-play-singletask-task2-v0$^{*}$} & & $9$ {\tiny $\pm 3$} & $\mathbf{91}$ {\tiny $\pm 4$} & $\mathbf{92}$ {\tiny $\pm 2$} & $0$ {\tiny $\pm 0$}& $\mathbf{95}$ {\tiny $\pm 2$}\\
         \tt{cube-triple-play-singletask-task3-v0} & 10M &  $12$ {\tiny $\pm 5$}   &  $0$ {\tiny $\pm 0$}  &  $\mathbf{92}$ {\tiny $\pm 2$} & $0$ {\tiny $\pm 0$}& $\mathbf{88}$ {\tiny $\pm 3$}\\
         \tt{cube-triple-play-singletask-task4-v0} & &  $0$ {\tiny $\pm 1$}   &  $0$ {\tiny $\pm 0$}  &  $\mathbf{59}$ {\tiny $\pm 7$} & $0$ {\tiny $\pm 0$}& 45 {\tiny $\pm 7$}\\
         \tt{cube-triple-play-singletask-task5-v0} & &  $2$ {\tiny $\pm 1$}   &  $0$ {\tiny $\pm 0$}  &  $74$ {\tiny $\pm 4$} & $0$ {\tiny $\pm 0$}& $\mathbf{87}$ {\tiny $\pm 5$}\\
         \midrule
         \tt{puzzle-4x4-play-singletask-task1-v0} & &  54 {\tiny $\pm 4$}   &  28 {\tiny $\pm 5 $}  &  $66$ {\tiny $\pm 17$} & $0$ {\tiny $\pm 0$}& $\mathbf{92}$ {\tiny $\pm 8$}\\
         \tt{puzzle-4x4-play-singletask-task2-v0} & &  24 {\tiny $\pm 3$}   &  2 {\tiny $\pm 1$}  & $\mathbf{80}$ {\tiny $\pm 16$} & $0$ {\tiny $\pm 0$}& $42$ {\tiny $\pm 7$}\\
         \tt{puzzle-4x4-play-singletask-task3-v0} & 10M &  36 {\tiny $\pm 4$}   &  42 {\tiny $\pm 7$}  & $69$ {\tiny $\pm 22$} & $0$ {\tiny $\pm 0$}& $\mathbf{99}$ {\tiny $\pm 1$}\\
         \tt{puzzle-4x4-play-singletask-task4-v0$^{*}$} & & $22$ {\tiny $\pm 2$} & $28$ {\tiny $\pm 3$} & $70$ {\tiny $\pm 17$} & $0$ {\tiny $\pm 0$}& $\mathbf{88}$ {\tiny $\pm 4$}\\
         \tt{puzzle-4x4-play-singletask-task5-v0} & &  $22$ {\tiny $\pm 4$}   &  $3${\tiny $\pm 2$}  &  $61${\tiny $\pm 19$} & $0$ {\tiny $\pm 0$}& $\mathbf{89}$ {\tiny $\pm 6$}\\
         \midrule
         \tt{cube-quadruple-play-singletask-task1-v0} & &  $79$ {\tiny $\pm 6$}   &  $70$ {\tiny $\pm 9$}  &  $79$ {\tiny $\pm  7$} & $0$ {\tiny $\pm 0$}& $\mathbf{92}$ {\tiny $\pm 5$}\\
         \tt{cube-quadruple-play-singletask-task2-v0$^{*}$} & & $0$ {\tiny $\pm 0$} & $\mathbf{97}$ {\tiny $\pm 2$} & $63$ {\tiny $\pm 7$} & $0$ {\tiny $\pm 0$}& $\mathbf{100}$ {\tiny $\pm 0$}\\
         \tt{cube-quadruple-play-singletask-task3-v0} & 100M &  $6${\tiny $\pm 3$}   &  1 {\tiny $\pm 1$}  &  $33$ {\tiny $\pm 7$} & $0$ {\tiny $\pm 0$}& $\mathbf{62}$ {\tiny $\pm 9$}\\
         \tt{cube-quadruple-play-singletask-task4-v0} & & $0$ {\tiny $\pm 0$}   &  $13$ {\tiny $\pm 5$}  &  $\mathbf{38}$ {\tiny $\pm 7$} & $0$ {\tiny $\pm 0$}& 31 {\tiny $\pm 7$}\\
         \tt{cube-quadruple-play-singletask-task5-v0} & & 0 {\tiny $\pm 0$}   &  $0$ {\tiny $\pm 0$}  & $12$ {\tiny $\pm 6$} & $0$ {\tiny $\pm 0$}& $\mathbf{35}$ {\tiny $\pm 10$}\\
         \bottomrule
    \end{tabular}
    }
    \vspace{-0.1in}
    \label{tab:full}
\end{table}

\end{document}